\documentclass[10pt,journal,compsoc]{IEEEtran}
\usepackage{amsfonts}
\usepackage{amsmath}
\usepackage{amssymb}
\usepackage{bm}
\usepackage{cite}
\usepackage{graphicx}
\usepackage{epstopdf}
\usepackage{epsfig}
\usepackage{url}
\usepackage{setspace}
\usepackage{subfigure}
\usepackage{color}
\usepackage{algorithmicx}
\usepackage{algpseudocode}
\usepackage{algorithm}
\usepackage{threeparttable}
\usepackage{lineno}
\usepackage{multirow}
\usepackage{verbatim}
\usepackage{booktabs}
\graphicspath{{./figures/}}

\begin{document}

\title{Underground Diagnosis Based on GPR and Learning in the Model Space}

\author{Ao Chen, Xiren Zhou, Yizhan Fan,  Huanhuan~Chen,~\IEEEmembership{Senior Member,~IEEE}
	
\IEEEcompsocitemizethanks{\IEEEcompsocthanksitem Ao Chen, Xiren Zhou, Yizhan Fan, and Huanhuan Chen are with School of Computer Science and Technology, University of Science and Technology of China, Hefei, 230027, China; E-mail: chenao57@mail.ustc.edu.cn, zhou0612@ustc.edu.cn, fyz666@mail.ustc.edu.cn, hchen@ustc.edu.cn. Corresponding Author: Huanhuan Chen, Xiren Zhou.}
}

\IEEEtitleabstractindextext{
\begin{abstract}
Ground Penetrating Radar (GPR) has been widely used in pipeline detection and underground diagnosis. 
In practical applications, the characteristics of the GPR data of the detected area and the likely underground anomalous structures could be rarely acknowledged before fully analyzing the obtained GPR data, causing challenges to identify the underground structures or abnormals automatically. 
In this paper, a GPR B-scan image diagnosis method based on learning in the model space is proposed.
The idea of learning in the model space is to use models fitted on parts of data as more stable and parsimonious representations of the data.
For the GPR image, 2-Direction Echo State Network (2D-ESN) is proposed to fit the image segments through the next item prediction. 
By building the connections between the points on the image in both the horizontal and vertical directions, the 2D-ESN regards the GPR image segment as a whole and could effectively capture the dynamic characteristics of the GPR image. 
And then, semi-supervised and supervised learning methods could be further implemented on the 2D-ESN models for underground diagnosis.
Experiments on real-world datasets are conducted, and the results demonstrate the effectiveness of the proposed model. 
\end{abstract}

}

\maketitle

\IEEEdisplaynontitleabstractindextext

\IEEEpeerreviewmaketitle

\section{Introduction}
\IEEEPARstart{M}{odern} cities are facilitated by a large number of urban roads and subsurface facilities. Unlike above-ground buildings, the health of some urban roads and facilities requires exploration of the medium beneath the ground to be effectively estimated. Without proper diagnosis and maintenance, some aging roads or facilities might suffer from various failure modes, bringing urban hazards such as moisture damage, land subsidence, and infrastructure collapse\cite{metje2007mapping}. As one of the most suitable means for imaging the subsurface, Ground Penetrating Radar (GPR) makes the use of the transmission and reflection of the electromagnetic (EM) waves to detect dielectric properties changes in host materials\cite{daniels2004ground, hao2020air}. 

Considerable efforts have been devoted to interpreting GPR B-scan images, which could be roughly divided into two categories: identifying and fitting hyperbolic characteristics on B-scan images and detecting non-hyperbolic features. 
The hyperbolic characteristic in the B-scan image is generated by linear cylinder objects where GPR has moved across. The radius and depth of the object could be estimated by identifying and fitting the hyperbolic characteristic\cite{shihab2005radius}. 
Graphic methods \cite{porrill1990fitting,borgioli2008detection,windsor2014data}, machine learning methods \cite{youn2002automatic,maas2013using,pasolli2009automatic}, and some methods that combine multiple approaches\cite{chen2010probabilistic,chen2010robust,dou2016real,zhou2018automatic} have been employed to extract and fit hyperbolic features from noisy GPR B-scan images.

Besides hyperbolic characteristics generated by linear cylinder objects, non-hyperbolic features might be more common and realistic targets in GPR B-scan images when imaging the subsurface. These features could be formed by different kinds of subsurface media or targets, including subsurface cavities, moisture damage, loose media, etc. Different subsurface media or targets would generate different characteristics of the image. Even the same kind of underground target could form different detail features on B-scan images due to differences in its composition, size, and surrounding media\cite{jol2008ground}.
For example, the moisture-damaged area would form as a continuous or discontinuous highlight area in GPR B-scan images, while the water content and whether the surrounding is hollow will feasibly affect the bright changes and edges of the area in the obtained B-scan images\cite{ferrara2015detecting}. Some dielectric-constant-based methods \cite{hammons2006detection, plati2013estimation} and signal processing algorithms \cite{liu20203d,rasol2020gpr} have been proposed to localize the subsurface objects. Recently, Convolutional Neural Network (CNN) methods\cite{tong2018innovative,tong2020pavement,ukhwah2019asphalt,du2021pavement,liu2021application,cong2021rrnet,hou2022neural} have also been utilized in recognizing objects on the image.

\begin{figure}[htpb]
	\centering
	\includegraphics[width=0.35\textwidth]{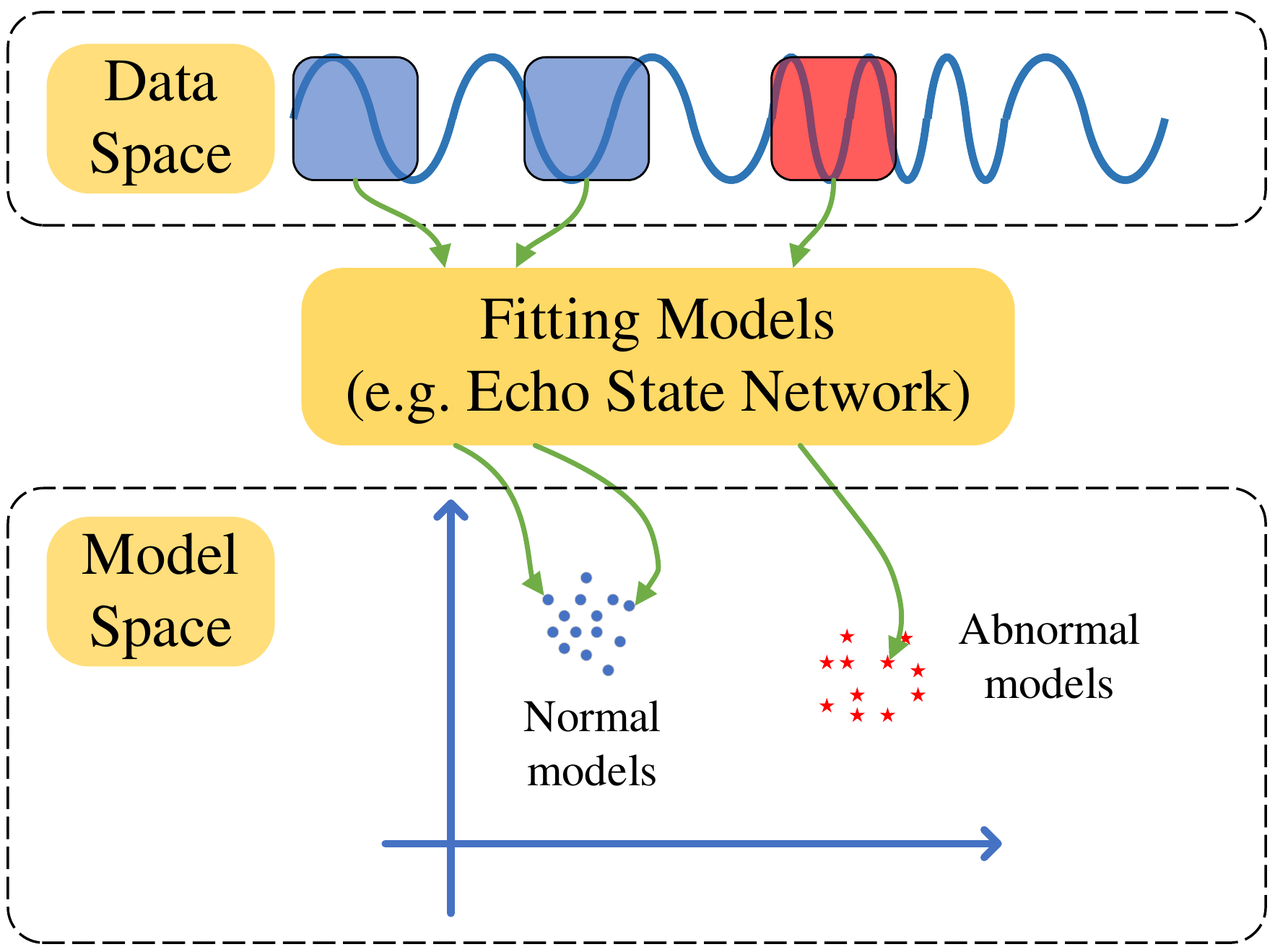}
	\caption{Illustration of “learning in the model space” framework. The model space is constructed by fitting a series of models using a series of data segments selected with a sliding window. Learning methods could be further implemented in the model space instead of in the data space.}
	\label{modelspace}
\end{figure}

\begin{figure*}[t]
	\centering
	\includegraphics[width=0.96\textwidth]{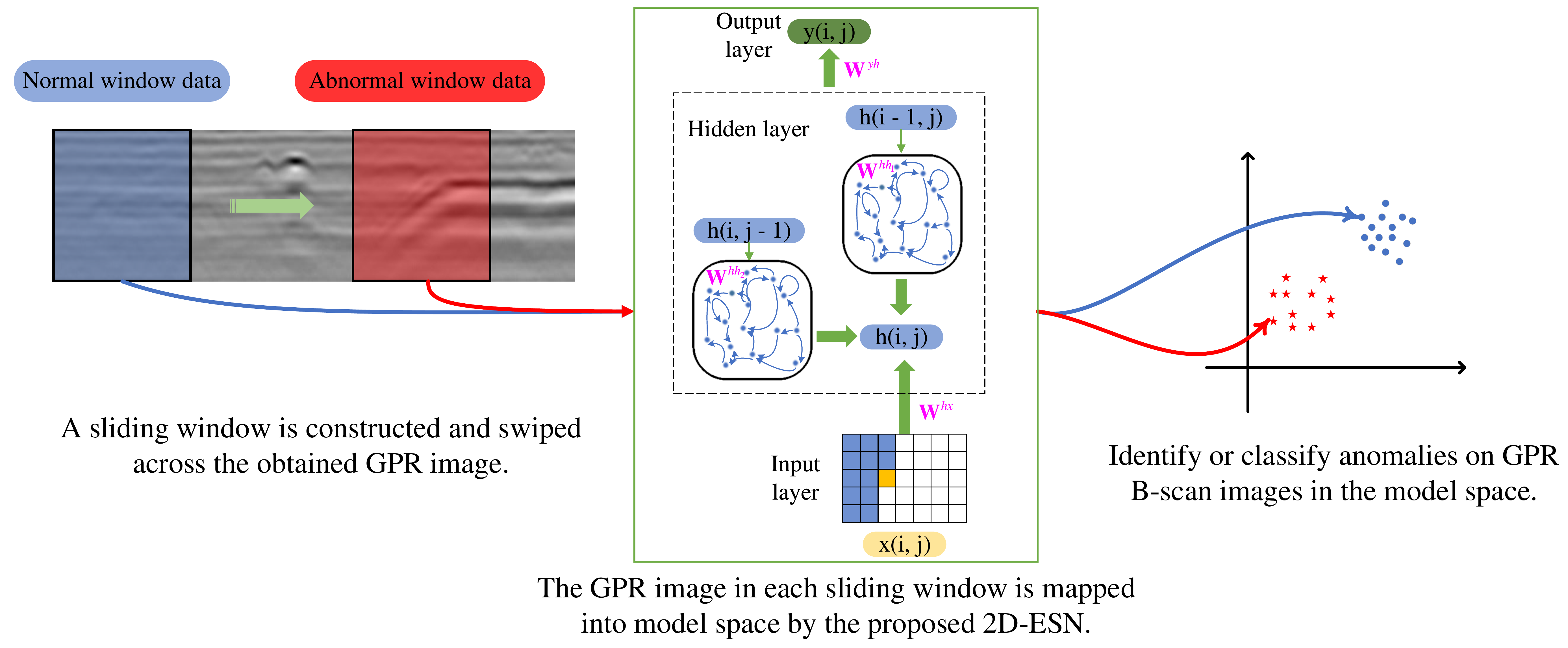}
	\caption{The procedure of the proposed method. A sliding window is constructed and swiped across the obtained GPR image. Among them, the normal window data means that there is no abnormal underground structure in the sliding window, and the abnormal window data means that there exist abnormal underground structures in the sliding window. The GPR image in each sliding window is mapped into the model space by the proposed 2D-ESN. Supervised and semi-supervised algorithms are then utilized to identify or classify anomalies on GPR B-scan images in the model space.}
	\label{Procedure}
\end{figure*}

In order to identify anomalies in the data, model-space-based methods have been proposed and used for data classification and diagnosis\cite{chen2013learning,quevedo2014combining,chen2014cognitive,chen2015model,gong2018multiobjective}. 
As shown in Fig. \ref{modelspace}, the model-space-based methods map the data from the data space to the model space by fitting models and use the models that could describe the dynamic characteristics\footnote{Dynamic characteristics refer to the changing and fluctuating laws of the data with continuity. Especially, in this paper, dynamic characteristics refer to the changing laws of the gray value of the GPR B-scan image.} of the data to represent the data or data clusters. In \cite{chen2013learning}, temporal signal data was segmented and mapped from signal space to model space. The Cycle topology with Regular Jumps (CRJ)\cite{rodan2010minimum,rodan2012simple} was utilized to fit the signal segments. The fitted CRJ model for each data segment was further classified in the model space. Subsequently, in \cite{quevedo2014combining} and \cite{chen2014cognitive}, the method of learning in the model space was utilized in the fault diagnosis of the Barcelona water network and the Tennessee Eastman Process. In \cite{chen2015model} and \cite{gong2018multiobjective}, time-series data was mapped into the model space and then classified. The above model-space-based methods aim to handle data with contextual relationships. However, the GPR B-scan image not only has the continuity of detection time or position in the horizontal direction but also correlates in the vertical direction due to the continuity of the underground medium.
The fitting models used in the above model-space-based methods are constructed on the traditional Echo State Network (ESN)\cite{jaeger2001echo}, which might not be able to capture the dynamics of two-dimensional image data.
Therefore, there is a need for a designed model that can fit or describe two-dimensional image data\cite{graves2007multi} to capture the dynamic characteristics of the GPR B-scan image and map the image into the model space with a proper size for further processing.

In this paper, a GPR B-scan image diagnosis method based on learning in the model space is proposed. A sliding window is constructed and swiped across the obtained GPR image. The GPR image segments in the sliding window are mapped into the model space with the proposed 2-Direction Echo State Network (2D-ESN).
Specifically, the 2D-ESN is a model that considers both current information and historical information in two different directions.
In GPR images, each point is correlated with surrounding points in the horizontal and vertical directions.
Different underground structures will show different variation laws in the horizontal and vertical directions in the GPR B-scan image, that is, the dynamic characteristics. 
By constructing the connections between the points on the image in both the horizontal and vertical directions, the 2D-ESN could effectively capture the dynamic characteristics of the GPR image.
As a result, GPR image segments of similar structures would be mapped to similar 2D-ESN models. On the contrary, the 2D-ESN models fitted from GPR image segments generated by different underground structures would have large differences in the model space due to the different dynamic characteristics. 
Moreover, the 2D-ESN is a model that could fit B-scan images without training, enabling real-time underground diagnosis.
Subsequently, the distance measuring method for 2D-ESN models is modified in the constructed model space. Based on the constructed model space and distance measuring method, supervised and semi-supervised algorithms could be utilized to identify or classify anomalies on GPR B-scan images in the model space. The procedure of the proposed method is visualized in Fig. \ref{Procedure}.
The main contribution of this paper could be summarized as follows:

\begin{enumerate}
	\item In the proposed 2D-ESN, both detection position and time continuity in the horizontal direction and medium continuity in the vertical direction of the GPR B-scan images are taken into account to capture the dynamic characteristics of the B-scan image effectively.
	\item For an image with the depth (length of a column) of $M$, the proposed 2D-ESN with $N$ hidden layer units could fit a model of size $2N$ instead of an $M\times N$ model obtained by the traditional ESN, which could effectively reduce the memory usage and enable the proposed diagnosis method to run efficiently on a personal computer for real-time processing.
	\item The proposed diagnosis method could perform real-time diagnosis of GPR images without sufficient prior knowledge and data and further classify different types of abnormal data in the constructed model space.
\end{enumerate}

The rest of this paper is organized as follows. Some background work is presented in Section II. The 2D-ESN is discussed in Section III, by which the GPR B-scan image segment could be mapped into the model space. Section IV provides the diagnosis in GPR images based on learning in the model space. Experiments are conducted and analyzed in Section V. Finally, conclusions are drawn in Section VI.

\section{Background}

\begin{figure}[htbp]
	\centering
	\includegraphics[width=0.35\textwidth]{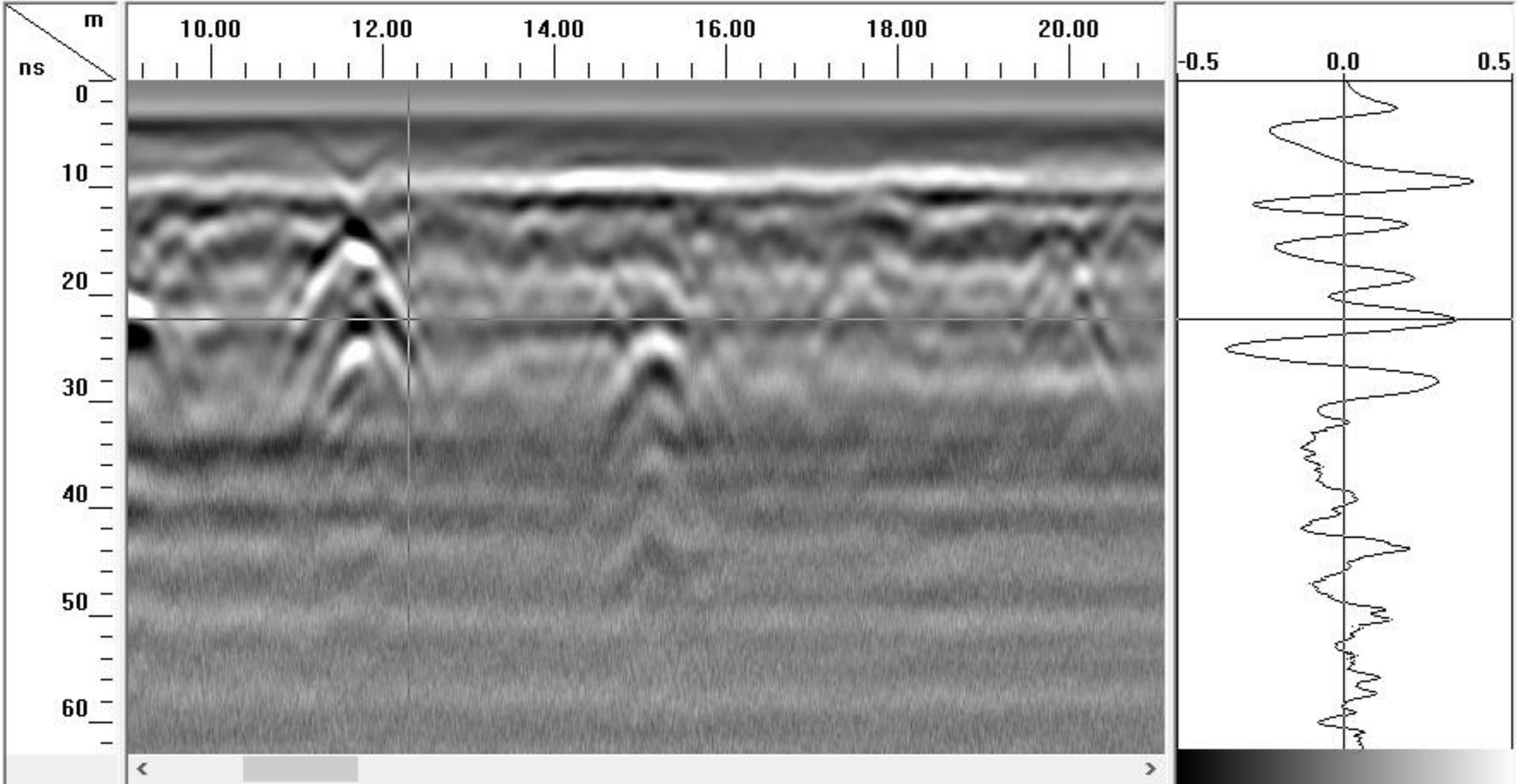}
	{\caption{This figure shows the generation of a B-can image. The left side of the figure is a B-scan image. The abscissa in the left B-scan picture is the detection position, and the ordinate is the time (the time from emission to reception of the electromagnetic wave). The right side is a received wave in the left B-scan image. The bottom of the right side shows that different gray values correspond to different magnitudes. The gray value of each pixel represents the corresponding amplitude at this position and time.}
		\label{abscan}}
\end{figure}

Ground Penetrating Radar (GPR) has been widely used as a non-destructive tool to image the subsurface\cite{daniels2004ground}. The GPR transmitter and antenna emit electromagnetic energy into the ground. When the energy encounters a buried object or a boundary between materials having different permittivities, it may be reflected or scattered back to the surface. A receiving antenna would then record the variations in the return waves. 
As Fig. \ref{abscan} shows, a GPR B-scan image could be obtained by arranging the received waves horizontally according to the time or space relationship and using the corresponding gray value to represent the wave intensity \cite{hao2020air}.
Published methods for interpreting GPR B-scan images could be roughly divided into two categories: identifying and fitting hyperbolic features on B-scan images and detecting non-hyperbolic features. 
The hyperbolic features in the B-scan image are generated by linear cylinder objects where GPR has moved across. In practical applications, the obtained B-scan images could be noisy. Some
graphic methods \cite{porrill1990fitting,borgioli2008detection,windsor2014data} and machine learning methods \cite{youn2002automatic,maas2013using,pasolli2009automatic} have been proposed to identify hyperbolic shapes on the GPR image. 
Combinations of the above methods \cite{chen2010probabilistic,chen2010robust,dou2016real,zhou2018automatic} have also been utilized to obtain more accurate and reliable results.  

Apart from hyperbolic features, non-hyperbolic features could be formed by different kinds of subsurface media or targets, including subsurface cavities, moisture damage, loose media, etc.
Some existing studies relied on the dielectric constant method with the air-coupled antenna to localize the subsurface objects \cite{hammons2006detection, plati2013estimation}, where the dielectric constant value was assumed to be the same along the vertical direction within the road layer and calculated based on the reflectance amplitude of the GPR wave\cite{maser1991use}. However, the dielectric constant could be sensitive and easily affected by material constituents, air void content, pavement density, and layer thickness\cite{evans2012assessing}. Thus the use of this kind of method generally requires knowledge of the basic conditions of the underground medium in advance, and its effect could be affected by the user's experience.
There are also some published works that identify underground objects from GPR data by signal processing or image recognition methods. In \cite{liu20203d}, the frequency-domain-focusing (FDF) technology of synthetic aperture radar (SAR) was utilized to aggregate scattered GPR signals for acquiring testing images, where a low-pass filter was designed to denoise primordial signals, and the profiles of detecting objects were extracted via the edge detection technique based on the background information. Subsequently, in \cite{rasol2020gpr}, a formula was conducted to relate the hidden crack width with the relative measured amplitude. Likewise, methods based on data or signal characteristics generally require prior knowledge or examples of subsurface conditions or structures.

Recently, Convolutional Neural Network (CNN) methods have been utilized in recognizing objects in GPR B-scan images. Tong \emph{et al.} \cite{tong2018innovative,tong2020pavement} conducted a CNN structure to automatically localize several kinds of targets in GPR data, which used the GPR signals as an input value to import into the CNNs.In \cite{ukhwah2019asphalt,du2021pavement}, YOLO(You-Only-Look-Once)\cite{redmon2018yolov3} was utilized to detect potholes and crackings beneath the roads. In \cite{zhang2020automatic}, a mixed deep CNN model combined with the Resnet-based network was proposed to detect the moisture damage in GPR data. Liu \emph{et al.} \cite{liu2021application} proposed a method for combining the YOLO series with GPR images to recognize the internal defects in asphalt pavement. In real-world applications, when detecting the underground objects in a certain area or along a road, it is difficult to ensure that the existing training B-scan images obtained from other areas or datasets are consistent with the underground situation in the area or road to be detected. Moreover, collecting enough training data and training a suitable CNN for the detecting area or road could be time-consuming.

In this paper, a model-space-based approach for GPR B-scan image diagnosis is proposed. First, a sliding window is built, which slides over the obtained GPR images. The GPR images in each sliding window would then be mapped to the model space by the proposed 2D-ESN with the next item prediction task. Subsequently, the distance measurement method of the 2D-ESN models is modified in the constructed model space. Based on the above, supervised and semi-supervised algorithms could be utilized to identify or classify underground anomalies in the constructed model space.

\section{2-Direction Echo State Network}

In order to map the data into the model space, the model used to fit the data should adequately capture the dynamic characteristics of the data. Meanwhile, in the process of real-time diagnosis, newly collected data needs to be processed constantly, which also puts forward requirements for the solution and the memory occupation of the model. In this section, the 2-Direction Echo State Network (2D-ESN) is proposed to satisfy the above requirements. The Echo State Network (ESN) is briefly introduced in advance and the 2D-ESN is then detailed.

\subsection{A Brief Introduction to Echo State Network (ESN)}

A traditional ESN\cite{jaeger2001echo} could be considered as a recurrent discrete-time neural network that processes sequential data with context. To capture the dynamic characteristics of sequential data, the ESN takes into account not only the impact of current input but also the impact of historical information.
\begin{figure}[htpb]
	\centering
	\includegraphics[width=0.49\textwidth]{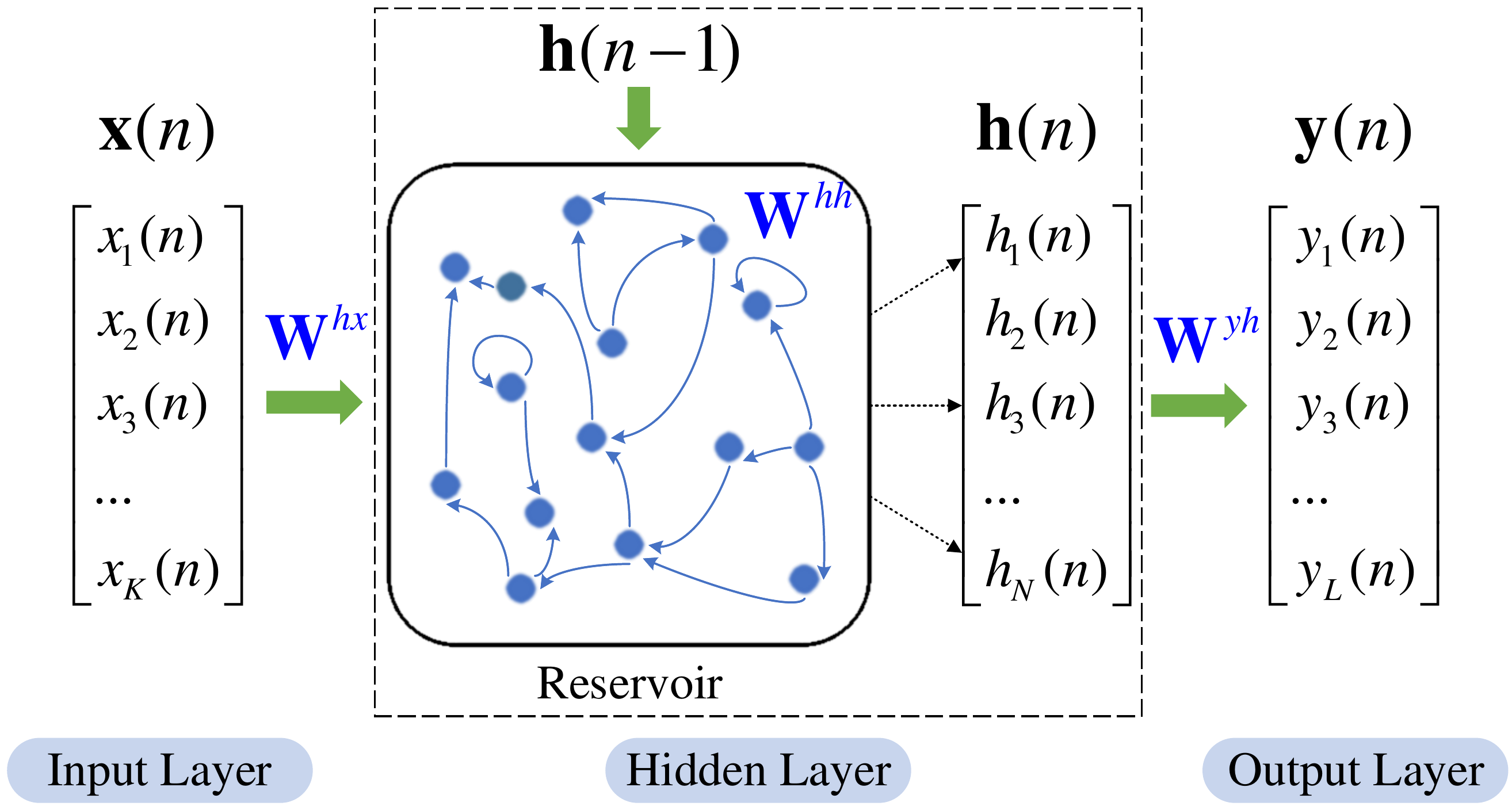}
	\caption{Schematic diagram of the traditional ESN. The input weight $\mathbf{W}^{hx}$ and the reservoir weight $\mathbf{W}^{hh}$ are randomly initialized and fixed before training. During the training process, the data enters the hidden layer step by step through the input layer in order. Then, the data is iterated over in the hidden layer to get the hidden state corresponding to each moment. Finally, the ridge regression is utilized to construct a hidden-to-output mapping to train the output weight $\mathbf{W}^{yh}$.}
	\label{ESN_procedure}
\end{figure}
As shown in Fig. \ref{ESN_procedure}, an ESN consists of an input layer with $K$ units, a hidden layer with $N$ units, and an output layer with $L$ units. The input value, hidden state, and output value at time step $n$ could be represented as $\mathbf{x}(n) = (x_1(n),..., x_K(n))^T$, $\mathbf{h}(n) = (h_1(n),...,h_N(n))^T$ and $\mathbf{y}(n) = (y_1(n),...,y_L(n))^T$, respectively. 
And the traditional ESN iteration formula is defined as:
\begin{equation}
	\label{eq1}
	\left\{
	\begin{aligned}
		\mathbf{h}(n) = &g(\mathbf{W}^{hh}\mathbf{h}(n-1) + \mathbf{W}^{hx}\mathbf{x}(n)), \\
		\mathbf{y}(n) = &\mathbf{W}^{yh}\mathbf{h}(n) + \mathbf{a},
	\end{aligned}
	\right.
\end{equation}
where  $\mathbf{W}^{hx}\in\mathbb{R}^{N\times K}$ is the weight between the input layer and the hidden layer, $\mathbf{W}^{hh}\in\mathbb{R}^{N\times N}$ is the reservoir\footnote{The reservoir is the component of the hidden layer and is utilized to process historical information.} weight in the hidden layer, $\mathbf{W}^{yh}\in\mathbb{R}^{L\times N}$ is the weight between the hidden layer and the output layer, $g$ is the activation function (typically $\tanh$), and $\mathbf{a}$ is the bias vector of the output model. 
In the iterative process of the hidden layer, the hidden state $\mathbf{h}(n)$ of the moment $n$ is affected by the current input value $\mathbf{x}(n)$ and the hidden state $\mathbf{h}(n-1)$ of the previous moment.

As a simplified Recurrent Neural Network (RNN), the input and reservoir weights in ESN are randomly generated and fixed before training. The reservoir in this kind of network should have the Echo State Property (ESP) \cite{jaeger2002tutorial}, which means that the reservoir would asymptotically wash out the effect of the historical information over time to ensure the stability of the ESN. The reservoir weight $\mathbf{W}^{hh}$ is scaled as:
\begin{equation}
	\mathbf{W}^{hh} \leftarrow \frac{ \alpha \mathbf{W}^{hh}} {\left\| \lambda_{max} \right\| },
\end{equation}
where $\left\| \lambda_{max} \right\|$ is the spectral radius\footnote{Spectral radius is the largest absolute value of an eigenvector. In the ESN, the effect of historical information on the current hidden state is related to the spectral radius of the reservoir weight. In this paper, the spectral radius of the reservoir weight is set from 0 to 1 to satisfy the ESP. It should be noted that this is not a necessary and sufficient condition for the ESP.} of $\mathbf{W}^{hh}$, and $0 < \alpha < 1$ is the scaling parameter.

When training an ESN model, we send the input data into the hidden layer iteratively, and the corresponding hidden states are then obtained. The output weights $\mathbf{W}^{yh}$ could be calculated by the ridge regression \cite{hoerl1970ridge} as:
\begin{equation}
	\label{eq2}
	\mathbf{W}^{yh} = (\mathbf{H}^T\mathbf{H} + \lambda^2\mathbf{I})^{-1}\mathbf{H}^T\mathbf{Y}, 
\end{equation}
where $\mathbf{I}$ is the identity matrix, $\mathbf{Y}$ is a vector of the target values, $\mathbf{H}$ is a vector of the corresponding hidden states, and $\lambda > 0$ is a regularization factor. 

With the ESN, sequential data with context could be mapped from the data space into the model space for further analysis. But there exist some limitations with ESN. The association of data in the ESN is unidirectional, while relationships in multiple directions within the data could not be captured. Moreover, the size of the ESN model increases linearly with the number of output units according to Equation \eqref{eq2}.
In GPR data processing, the input data could be the two-dimensional B-scan image.
Suppose there is a GPR image, and its vertical direction (depth) contains $M$ pixels. The size of the obtained ESN's output weights $\mathbf{W}^{yh}$ would be $M \times N$. Therefore, if the ESN is directly utilized for fitting image data, the parameter dimension of the output layer would be too large, which is not conducive to further analysis in the model space. 

\subsection{2-Direction Echo State Network (2D-ESN)}

As aforementioned, the GPR B-scan image not only has the continuity of detection time or position in the horizontal direction but also correlates with the vertical direction due to the continuity of the underground medium. Different underground structures will show different variation laws (dynamic characteristics) in the horizontal and vertical directions in the GPR B-scan image.
To capture the dynamic characteristics of an image, the 2D-ESN correlates each point in the image with the left and upper points and builds memory for the upper left points that have been processed. Moreover, the 2D-ESN reduces the number of parameters in the output layer by order of magnitude compared to the ESN methods used in existing model-space-based methods.

\begin{figure}[htpb]
	\centering
	\includegraphics[height=2.1in]{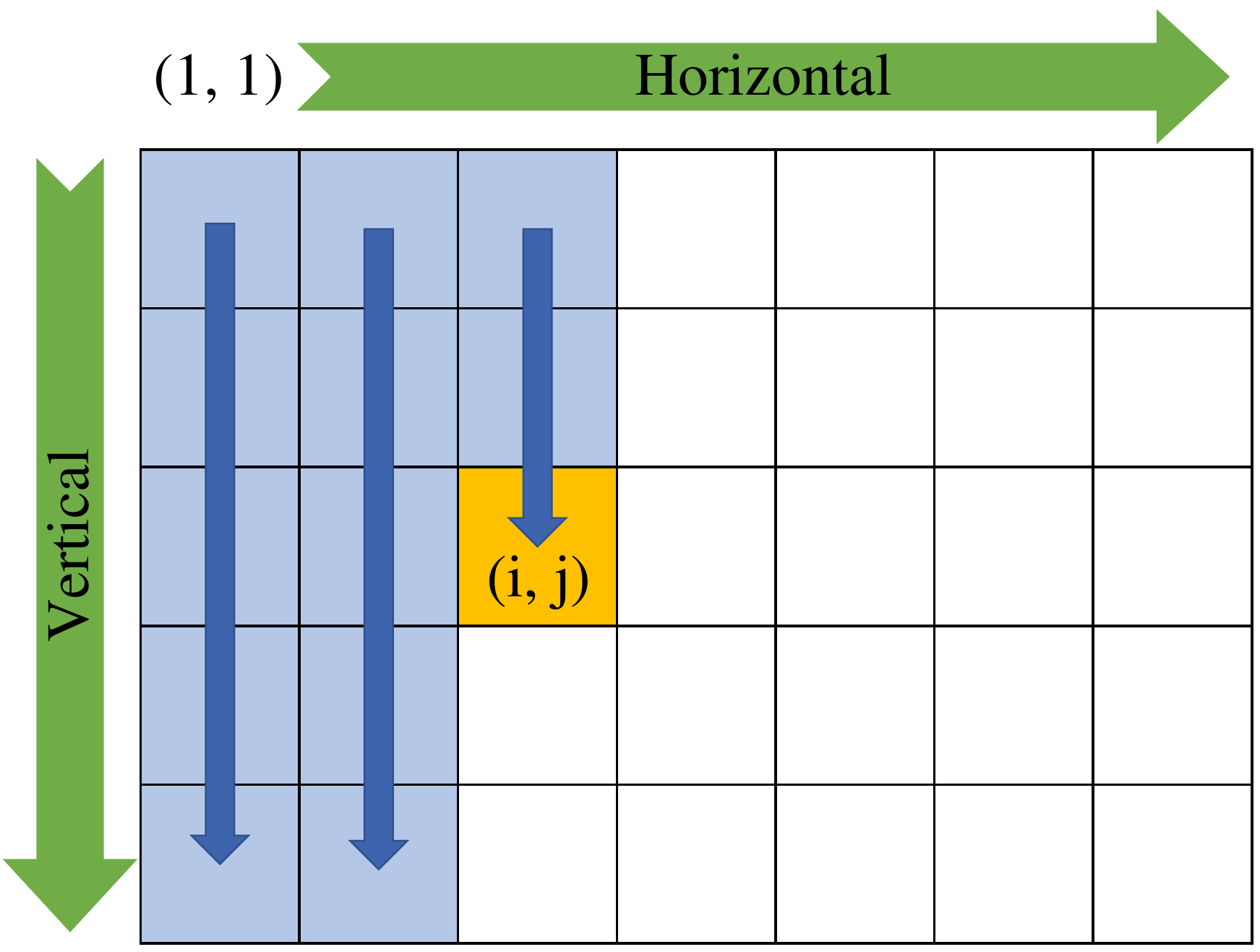}
	\caption{Schematic diagram of the processing flow of the 2D-ESN for image data. The 2D-ESN processes each point step by step along with the horizontal and vertical directions from the initial point $(1,1)$. Among them, blue parts indicate the points that have been processed and the yellow point $(i,j)$ is currently processing point.}
	\label{example}
\end{figure}

The processing flow of the 2D-ESN for image data is shown in Fig. \ref{example}. The 2D-ESN starts from the initial point $(1,1)$ in the image and gradually sends points into the reservoir for iteration from left to right and top to bottom. Similar to the ESN shown in Fig. \ref{ESN_procedure}, the 2D-ESN consists of the input layer, the hidden layer, and the output layer. The input layer indicates the gray value of the current point in the image. The output layer is the predicted value of the current point. The hidden layer combines the input and the hidden states of upper and left points to generate the hidden state of the current point. Among them, the input layer and hidden layer parameters do not need to be trained, and only the output layer parameters need to be solved by linear fitting.

\begin{figure}[htbp]
	\centering
	\subfigure[]{\centering
		\includegraphics[width=0.35\textwidth]{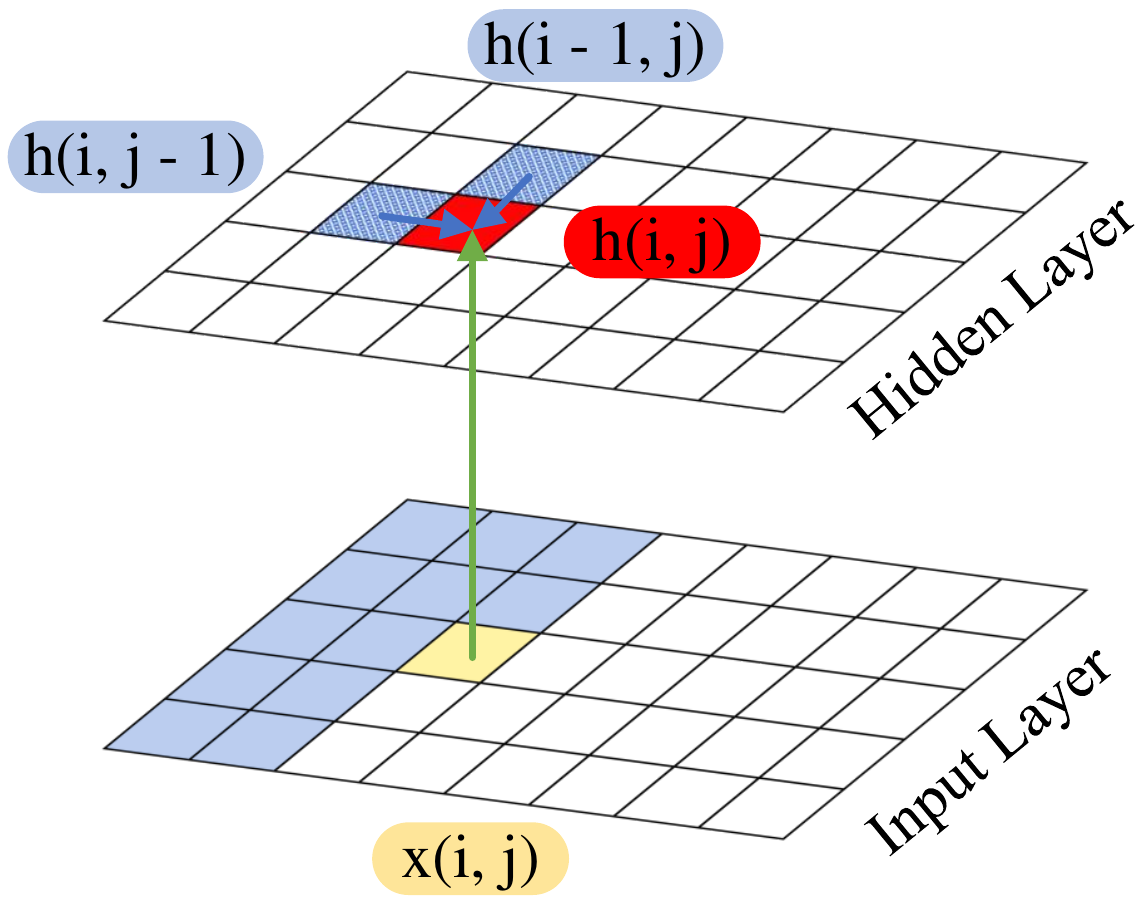}
		\label{2DESN_hidden}
	}
	\subfigure[]{\centering
		\includegraphics[width=0.30\textwidth]{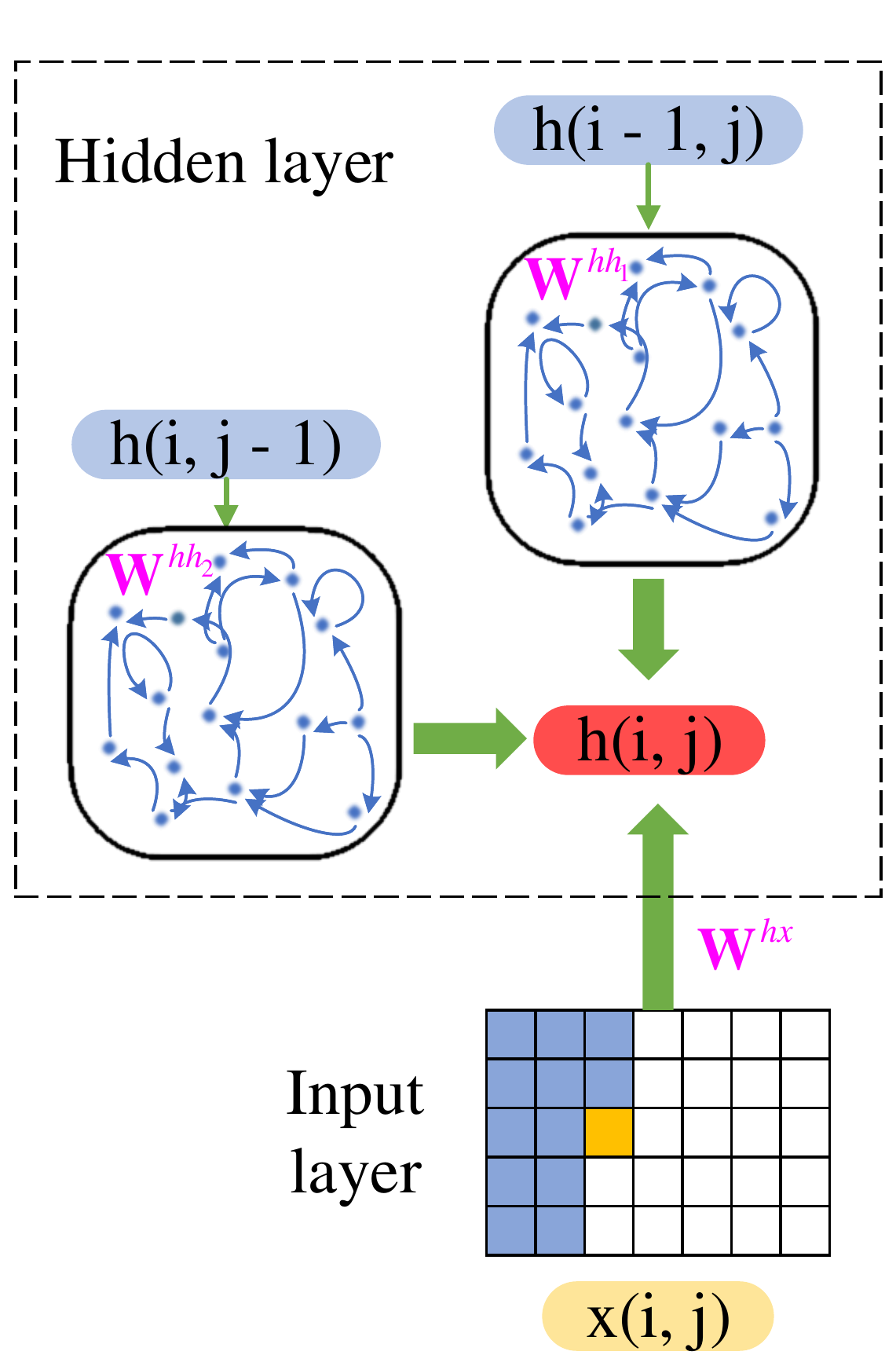}
		\label{2DESN_new}
	}
	\caption{(a) is the schematic diagram of the hidden layer iteration of the 2D-ESN. Hidden state $\mathbf{h}(i,j)$ is generated by the combination of gray value $\mathbf{x}(i,j)$ of the current point, hidden state $\mathbf{h}(i-1,j)$ of the upper point, and hidden state $\mathbf{h}(i,j-1)$ of the left point. (b) is the schematic diagram of the specific operation of Equation \eqref{eq3}.}
	\label{2DESN_itr}
\end{figure}

In the iteration procedure of the 2D-ESN, for each point, the current hidden state is not only influenced by the current input point but also by the hidden states of the upper and left points. For point $(i,j)$, the 2D-ESN hidden layer iteration is illustrated in Fig. \ref{2DESN_itr}, and the iteration formula is defined as:
\begin{equation}
	\label{eq3}
	\begin{split}
		\mathbf{h}(i,j) = &g(\mathbf{W}^{hh_1}\mathbf{h}(i-1,j) + \mathbf{W}^{hh_2}\mathbf{h}(i,j-1) +\\ &\mathbf{W}^{hx}\mathbf{x}(i,j)), 
	\end{split}
\end{equation}
where $\mathbf{h}(i,j)\in \mathbb{R}^{N\times1}$ is a hidden state of $(i,j)$ point in the image, $\mathbf{x}(i,j)\in \mathbb{R}^{1\times1}$ is the gray value of $(i,j)$ point in the image, $\mathbf{W}^{hh_1}$, $\mathbf{W}^{hh_2}\in \mathbb{R}^{N\times N}$ are the reservoir weights in the hidden layer, $\mathbf{W}^{hx}\in \mathbb{R}^{N\times1}$ is the weight of the input layer. Among them, the input weight ($\mathbf{W}^{hx}$) and reservoir weights ($\mathbf{W}^{hh_1}$, $\mathbf{W}^{hh_2}$) in the 2D-ESN are randomly generated. And the reservoir should satisfy the ESP. 

\begin{figure}[htpb]
	\centering
	\includegraphics[width=0.35\textwidth]{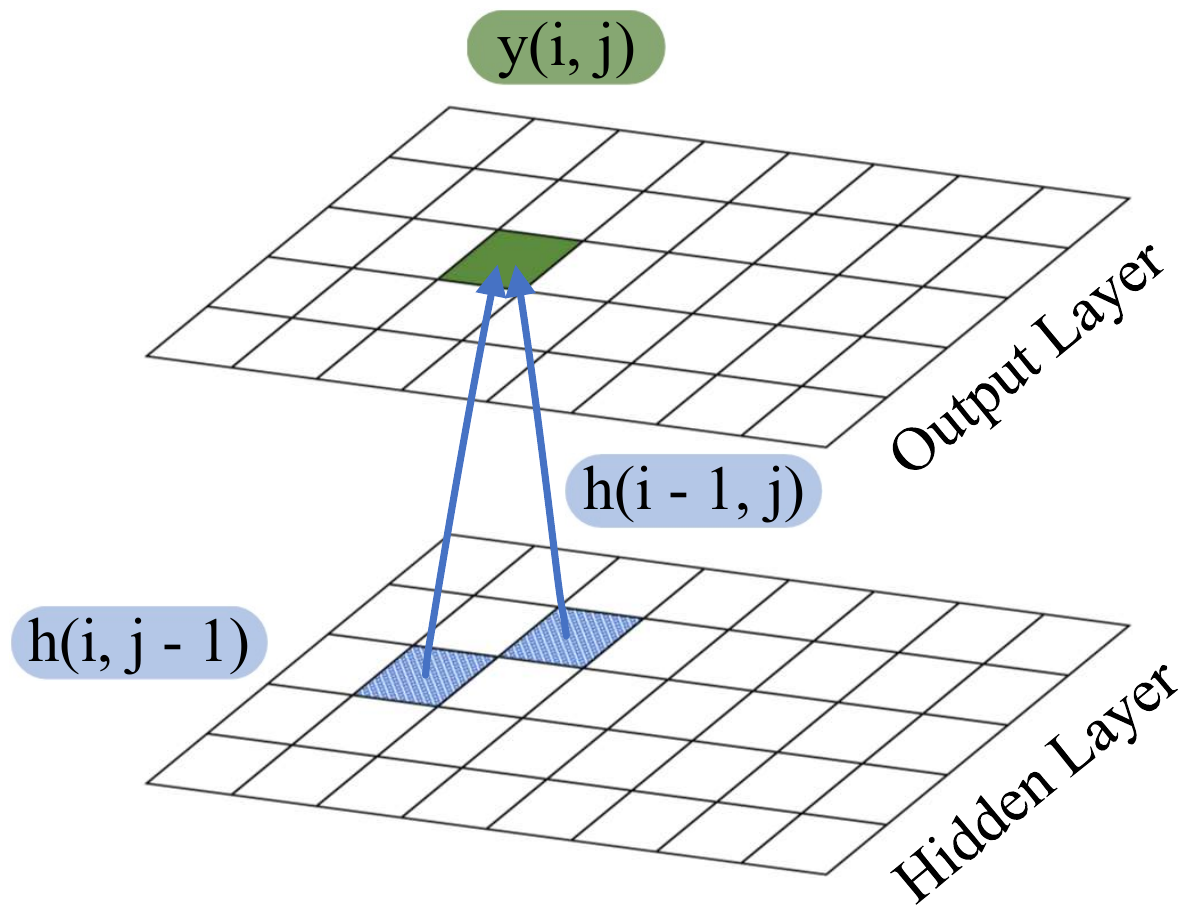}
	\caption{The schematic diagram of the next item prediction of the 2D-ESN. The predicted value $\mathbf{y}(i,j)$ of the current point is given by hidden state $\mathbf{h}(i-1,j)$ of the upper point and hidden state $\mathbf{h}(i,j-1)$ of the left point.}
	\label{2DESN_output}
\end{figure}

In order to map the image data into the model space through 2D-ESN, the next item prediction task is accomplished and shown in Fig. \ref{2DESN_output}, that is, to construct the mapping between the target value and its upper and left hidden states. For point $(i,j)$, the 2D-ESN prediction formula is defined as:
\begin{equation}
	\label{prediction}
	\mathbf{y}(i,j) = \mathbf{W}^{yh1}\mathbf{h}(i-1,j) + \mathbf{W}^{yh2}\mathbf{h}(i,j-1)) + \mathbf{a},
\end{equation}
where $\mathbf{y}(i,j)\in \mathbb{R}^{1\times1}$ is the output of $(i,j)$ point in the image, $\mathbf{h}(i-1,j), \mathbf{h}(i,j-1)\in \mathbb{R}^{N\times1}$ are the hidden state of upper and left point of current point $(i,j)$ in the image, and $\mathbf{a}$ is the bias vector of the output model. 
$\mathbf{W}^{yh} = [\mathbf{W}^{yh_1}, \mathbf{W}^{yh_2}]$ is the output weight of the 2D-ESN, where $\mathbf{W}^{yh_1}$, $\mathbf{W}^{yh_2}\in \mathbb{R}^{1\times N}$ could be obtained by the ridge regression as Equation \eqref{eq2}. According to the above prediction formula Equation \eqref{prediction} and the ridge regression Equation \eqref{eq2}, the 2D-ESN provides the output weight of size $2N$ instead of the output weight of size $M\times N$ provided by the ESN, where $M$ is the number of vertical pixels (depth) of the image.

When the 2D-ESN processes the point in the image with Equations \eqref{eq3} and \eqref{prediction} in the sequence shown in Fig. \ref{example}, it actually establishes a connection between the currently processed point and all the previous upper left points to capture the dynamic characteristics of the image horizontally and vertically. With the 2D-ESN, each segment of GPR B-scan images could be mapped from the data space to the model space by the next item prediction task.
The predictive model fitted through a GPR B-scan image segment indicates the mapped point of this image segment in the model space. Intuitively, differences between the fitted 2D-ESN models reflect differences in dynamic characteristics between the corresponding training data.
The method to measure the difference between the models in the model space would be introduced in the next section.

From the perspective of practical application requirements, real-time diagnostic results could aid in the localization and repair of subsurface anomalies. In the 2D-ESN, only $\mathbf{W}^{yh_1}$, $\mathbf{W}^{yh_2}\in \mathbb{R}^{1\times N}$ need to be solved by ridge regression when fitting a model, ensuring the speed and efficiency of the process of mapping data to the model space. Also, the memory occupied by the 2D-ESN is limited, which is helpful for performing real-time diagnostics on a personal computer in on-site applications.

\section{Diagnosis in GPR images based on model space}

In this section, the distance measuring method for the 2D-ESN is modified in the constructed model space. Based on the constructed model space and distance measuring method, learning methods in the model space, including semi-supervised and supervised algorithms, are introduced for underground diagnosis.

\subsection{Measure the Distance between Two 2D-ESNs}

After mapping GPR B-scan images to the model space via the 2D-ESN, the distance between models should be defined to measure the difference between models. The $m$-norm distance\cite{chen2013learning} between models $f_1(x)$ and $f_2(x)$ ($f_1, f_2:\mathbb{R}^{N\times2} \rightarrow \mathbb{R}$) could be defined as follows:
\begin{equation}
	\label{eq4}
	L_m(f_1,f_2) = (\int_C D_m(f_1(x),f_2(x))d\mu (x))^{1/m}, 
\end{equation}
where $D_m(f_1(x),f_2(x)) = \Vert f_1(x) - f_2(x) \Vert^m$ is a function that measures the difference between $f_1(x)$ and $f_2(x)$, $\mu(x)$ is the probability density function of the input domain $\mathbf{x}$, and $C$ is the integral range. In this paper, we adopt $m = 2$ and firstly assume that $\mathbf{x}$ is uniformly distributed. 

For two different 2D-ESN models, $f_1(x)$ and $f_2(x)$ could be represented by the following equation:
\begin{equation}
	\label{eq5}
	\left\{
	\begin{aligned}
		f_1(\mathbf{h}) = \mathbf{W}^{yh}_1\mathbf{h} + \mathbf{a}_1, \\
		f_2(\mathbf{h}) = \mathbf{W}^{yh}_2\mathbf{h} + \mathbf{a}_2, 
	\end{aligned}
	\right.
\end{equation}
where $\mathbf{h} = [\mathbf{h}(i-1,j), \mathbf{h}(i,j-1)]^T$ is the hidden states of the up and left points, $\mathbf{W}^{yh} = [\mathbf{W}^{yh_1}, \mathbf{W}^{yh_2}]$ is the weight of the output layer, and $\mathbf{a}$ is the bias vector of the output model.

Substituting Equation \eqref{eq5} into Equation \eqref{eq4}, the following equation could be obtained:
\begin{equation}
	\label{eq6}
	L_2(f_1,f_2) = \frac{1}{3}\Vert \mathbf{W}^{yh}_1 - \mathbf{W}^{yh}_2 \Vert^2 + \Vert \mathbf{a}_1 - \mathbf{a}_2 \Vert^2,
\end{equation}
where $L_2(f_1,f_2)$ is the obtained distance between two 2D-ESNs.
Through Equation \eqref{eq6}, the distance between any two 2D-ESNs could be measured. Thus the distance-based algorithms could be utilized in the model space for classification.

\subsection{Model Diagnostics in the Model Space}

In practical applications, the characteristics of the obtained GPR data of the detected area and the likely underground anomalous structures could be rarely acknowledged before analyzing the obtained GPR data. In this case, semi-supervised learning (one-class learning) is utilized for real-time underground diagnosis. In this paper, One-Class SVM (OCSVM)\cite{scholkopf2001estimating} is used to find a hyperplane that has the greatest distance from the origin in the kernel feature space with the given training data falling beyond the hyperplane. 
\begin{figure}[htbp]
	\centering
	\subfigure[]{ \centering
		\label{oneclass}
		\includegraphics[height=1.8in]{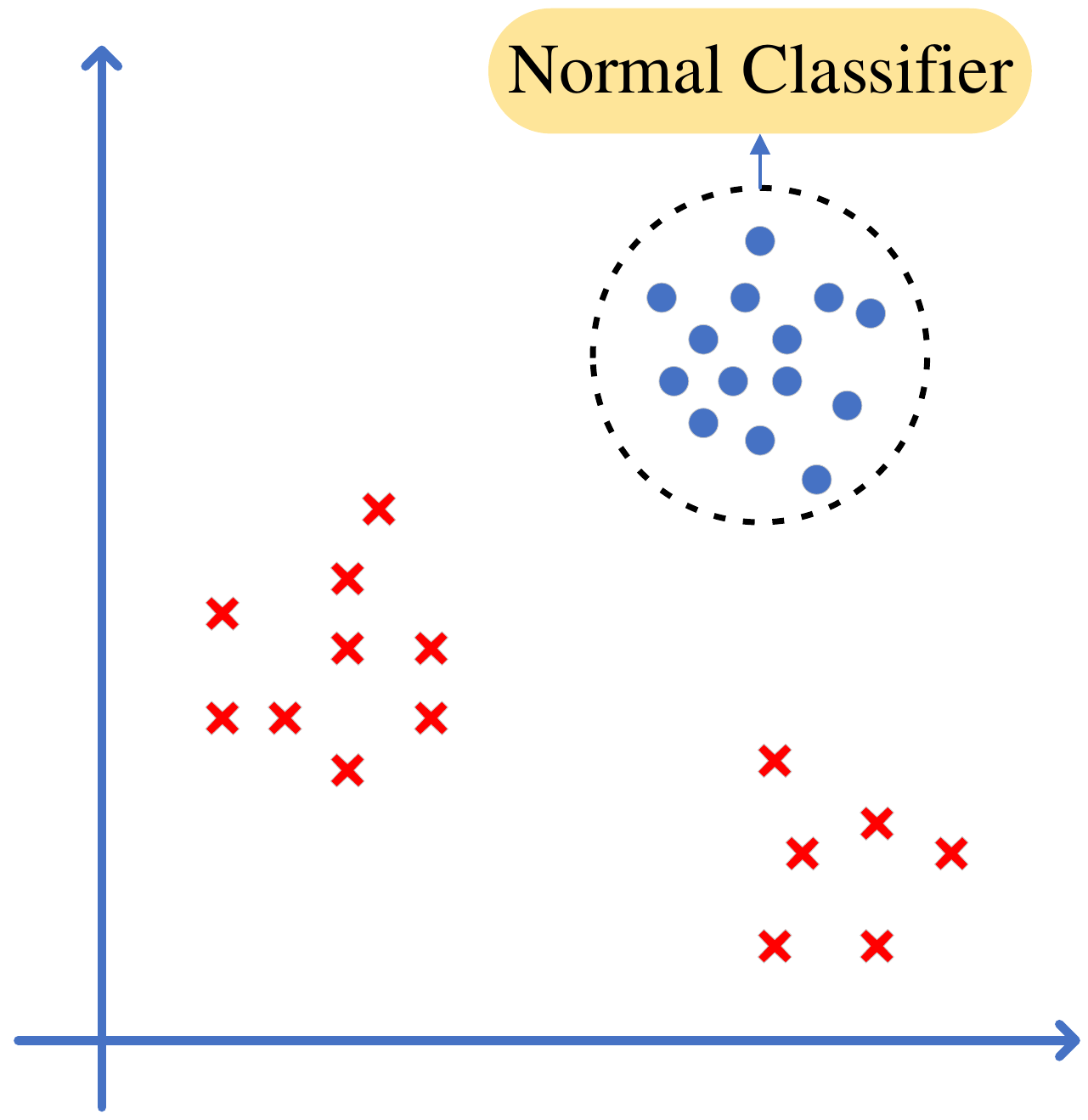}}\hspace{0.2in}
	\subfigure[]{ \centering
		\label{oneclass_inc}
		\includegraphics[height=1.8in]{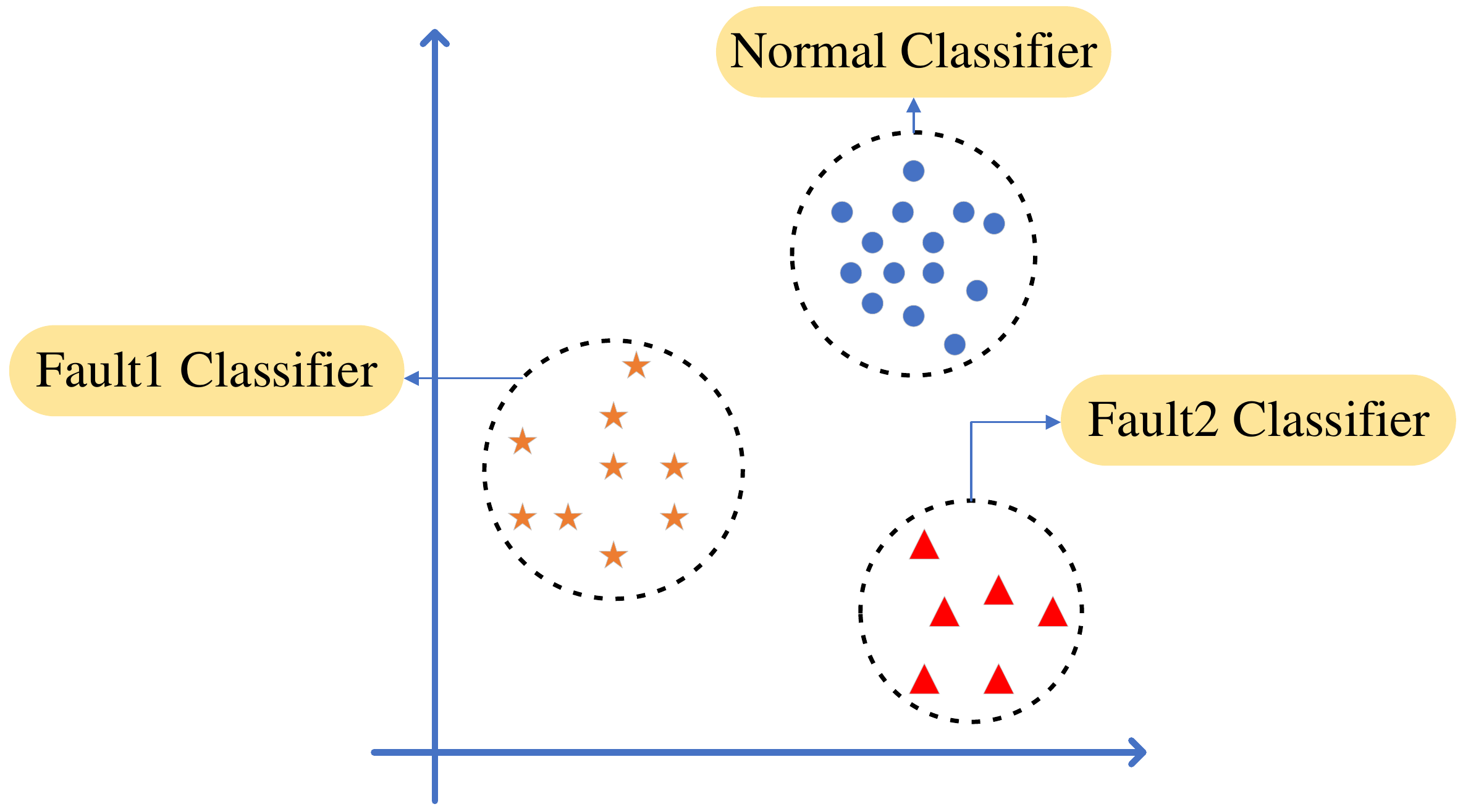}}
	\caption{(a) is the schematic diagram of one-class learning. Blue points are positive data, while red points are negative points. One-class learning only uses positive data to train a positive-negative classifier. (b) is the schematic diagram of incremental one-class learning. The points in the three circles represent the three types of data. And incremental one-class learning uses each one-class classifier to represent each normal/fault segment.}
	\label{ocleanring}
\end{figure}
Specifically, several normal GPR B-scan image segments are mapped to the model space via the 2D-ESN, and a normal OCSVM classifier is trained. Then, for the subsequent data, the OCSVM classifier trained from normal data is continuously used for classification, and the data classified as abnormal is put into the abnormal set. The abnormal data could be further trained by incremental one-class learning\cite{chen2013learning}. Fig. \ref{ocleanring} illustrates the schematic diagrams of one-class learning and incremental one-class learning.

For supervised classification tasks, that is when there is already data similar to the underground environment that can be used for model space mapping or training, the classification algorithms like K-Nearest Neighbors (KNN)\cite{altman1992introduction}, Random Forest \cite{ho1998random}, and Support Vector Machine (SVM)\cite{cortes1995support} could be utilized in the constructed model space.

\section{Experimental Study}

In this section, experiments on real-world datasets are conducted. After that, the analysis of the experimental results and some comparative works are presented.

\subsection{Experimental Settings on Real-World Datasets}

\begin{figure}[htbp]
	\centering
	\subfigure[]{ \centering
		\label{gprtest1}
		\includegraphics[height=1.125in]{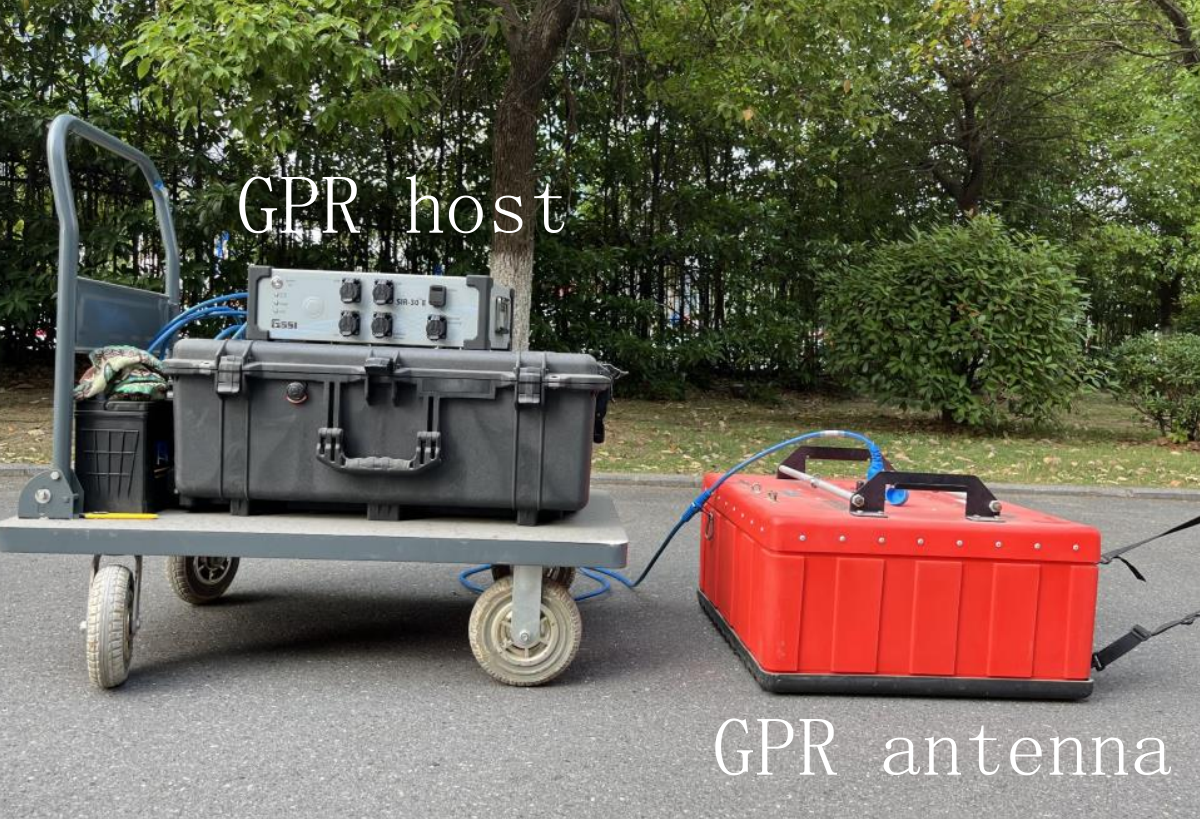}}
	\subfigure[]{ \centering
		\label{gprtest2}
		\includegraphics[height=1.125in]{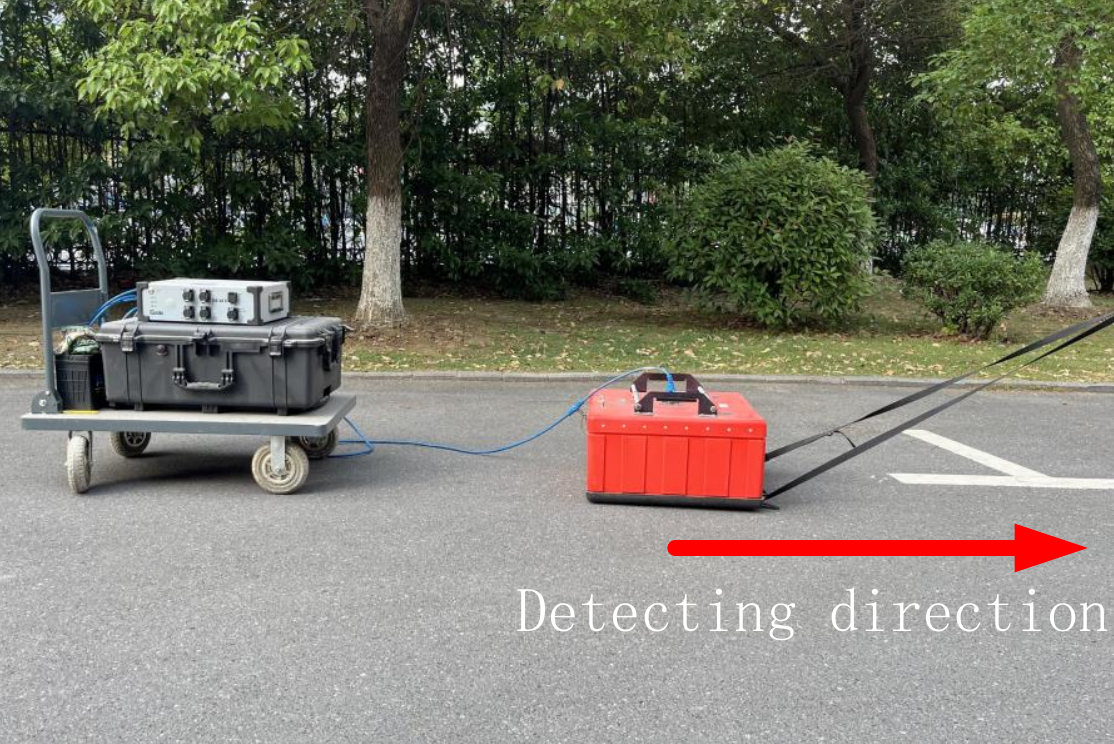}}
	\caption{(a) shows the host and antenna of the utilized GPR. (b) shows the actual detection scene of a  road. The GPR antenna is moved along the road (the red arrow) to obtain the GPR B-scan image, and the collected image is transformed to the host. As mentioned in the background, each column on the B-scan represents a received electromagnetic wave, and there is a sequence and connection between the columns in detection time or position.}
	\label{gprtest}
\end{figure}

To evaluate the effectiveness of the proposed model, experiments are conducted using detection data collected along 4 roads, including two concrete roads and two asphalt roads.
15 GPR B-scan images are obtained by the GSSI-SIR30 GPR along these roads. 
The utilized devices and detection are shown in Fig. \ref{gprtest}. 

The collected data has been manually classified, and some locations with part of underground anomalies were excavated and repaired. 
To eliminate the noise and highlight the subsurface structures, some operations are conducted on the obtained GPR B-scan images, which consist of three tasks: 1) Eliminating the undesired presence of the ground surface echo; 2) Reducing noise; 3) Compensating propagation losses. First, the reflectance of the ground surface is eliminated in advance. This work is supported by the Matgpr \cite{tzanis2006matgpr} which is a freeware Matlab package for the analysis of common-offset GPR data. Then a filtering step based on the standard median filter is performed \cite{olhoeft2000maximizing} to reduce the electromagnetic noise and interferences. Finally, concerning the compensation of the propagation losses caused by the medium attenuation and the signal energy radial dispersion, a nonlinear time-varying gain \cite{strange2002signal} is applied to the received signal. The above operations have been detailed in our previous work\cite{zhou2018automatic}. Thus, they are not expanded in this paper.

When mapping GPR data from the data space to the model space, the length of the sliding window is set to be 300 pixels (the distance between every two pixels is 0.141cm), and the sliding interval of the window is 20 pixels.
For the 2D-ESN model, since whether there is an abnormality at the current position will not be greatly affected by the long previous road segment, the spectral radius of the reservoir is set to 0.1, which means that the selected model has a small memory capacity and is more sensitive to data mutation. Before training, the input layer parameters and reservoir parameters of 2D-ESN are randomly generated and fixed.
After mapping each GPR B-scan image in the window, both the semi-supervised learning task and supervised learning task are conducted.

\subsection{Semi-Supervised and Supervised Learning Results in the Constructed Model Space}

For semi-supervised learning tasks, we select a part of the normal-road GPR B-scan image of length 3000 for each B-scan image obtained along the roads. And then, the sliding window is utilized on the image to generate a series of normal data segments, which is used to train a classifier by OCSVM. 
The remaining part of the road will be gradually segmented by the sliding window and sent into the 2D-ESN to obtain the corresponding model. Then, the model is classified as normal or abnormal by the trained classifier. Furthermore, if the road has anomalies that can be subdivided, incremental one-class learning is used to classify unknown anomalies. 

\begin{figure}[htpb]
	\centering
	\subfigure[]{ \centering
		\label{normal_example}
		\includegraphics[height=0.82in]{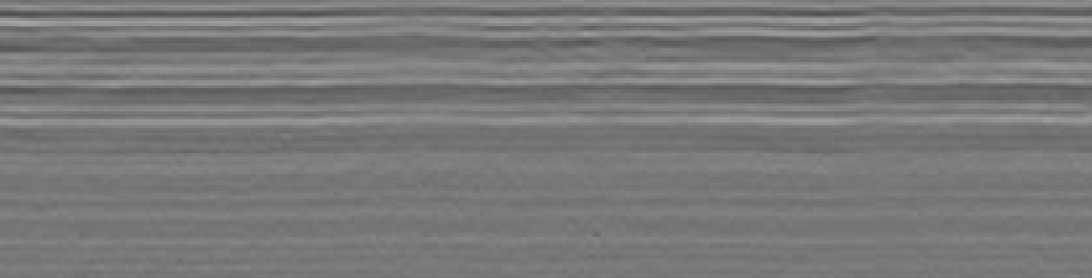}}
	\subfigure[]{ \centering
		\label{ab_example1}
		\includegraphics[height=0.71in]{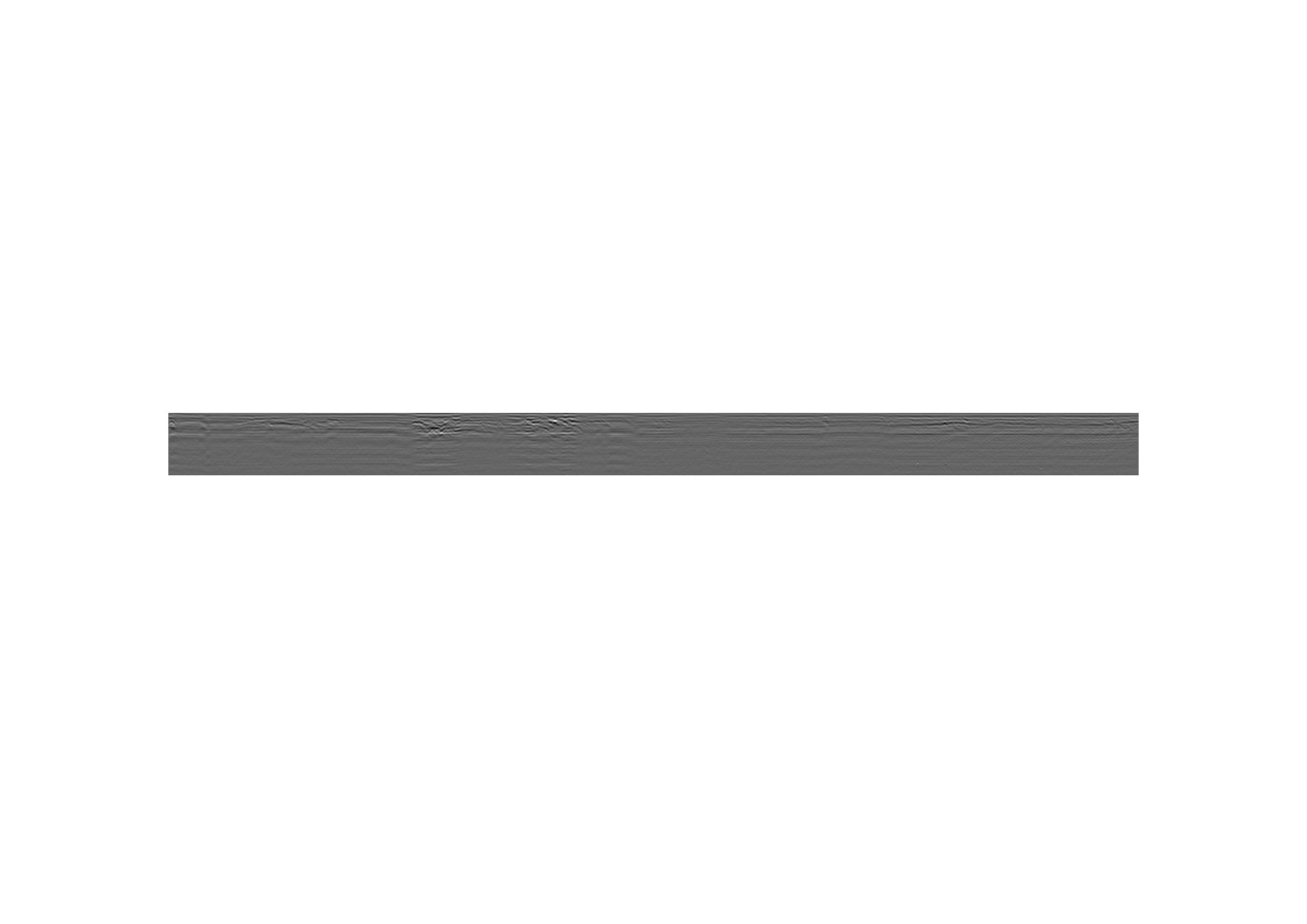}}
	\subfigure[]{ \centering
		\label{ab_example3}
		\includegraphics[height=0.71in]{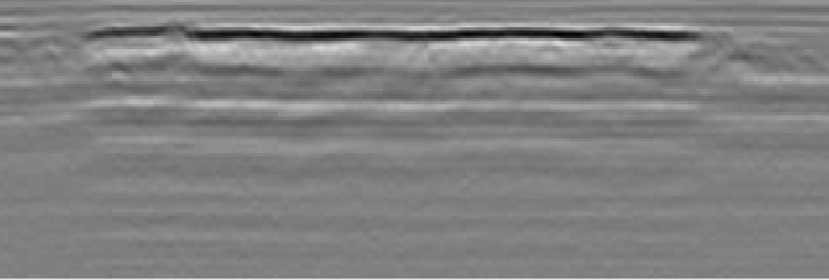}}	
	\subfigure[]{ \centering
		\label{ab_example2}
		\includegraphics[height=0.93in]{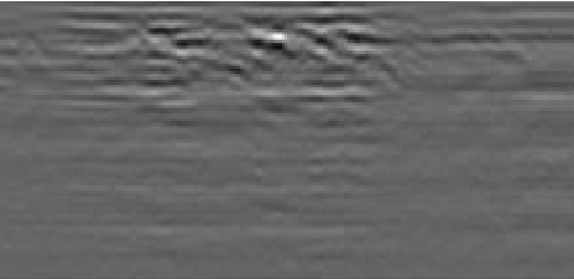}}	
	\subfigure[]{ \centering
		\label{ab_example4}
		\includegraphics[height=0.93in]{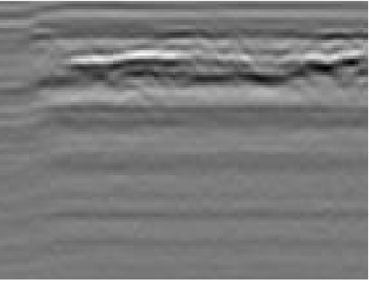}}
	\caption{(a) is the GPR B-scan image segment generated by a normal road without any underground structures. (b) to (e) illustrated some identified underground abnormals in the model space. After on-site inspection, it is determined that (b) was caused by loose subgrade, (c) was cracks, and (d) and (e) were rich in underground water. Since the length of the abnormal image would be longer than the sliding window, the overlapping GPR image segments of the same type are merged to obtain the final anomaly GPR image. }
	\label{abexample}
\end{figure}

\begin{figure*}[t]
	\centering
	\subfigure[]{ \centering
		\label{data32}
		\includegraphics[width=0.308\textwidth]{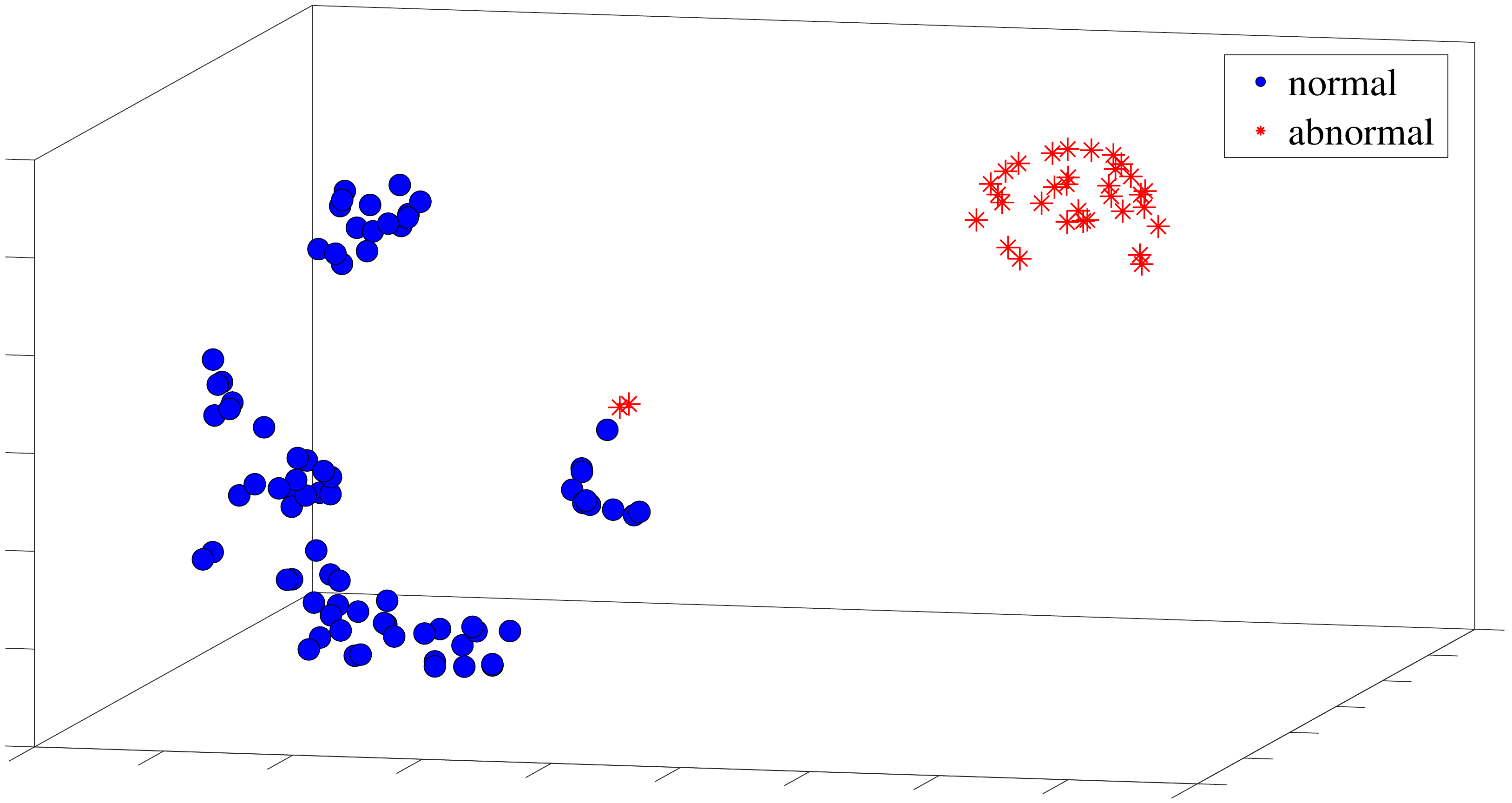}}
	\subfigure[]{ \centering
		\label{data92}
		\includegraphics[width=0.308\textwidth]{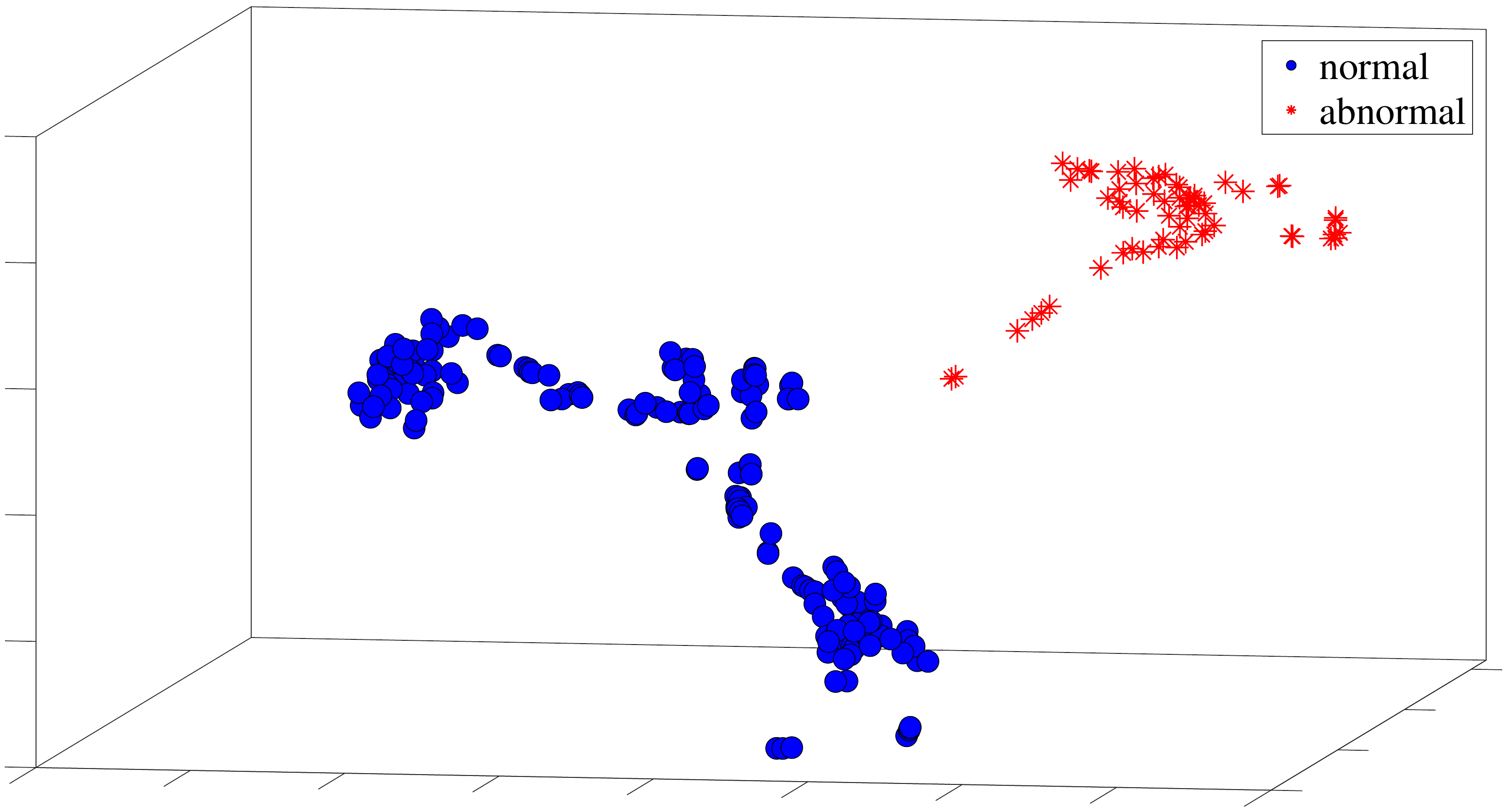}}	
	\subfigure[]{ \centering
		\label{data22}
		\includegraphics[width=0.308\textwidth]{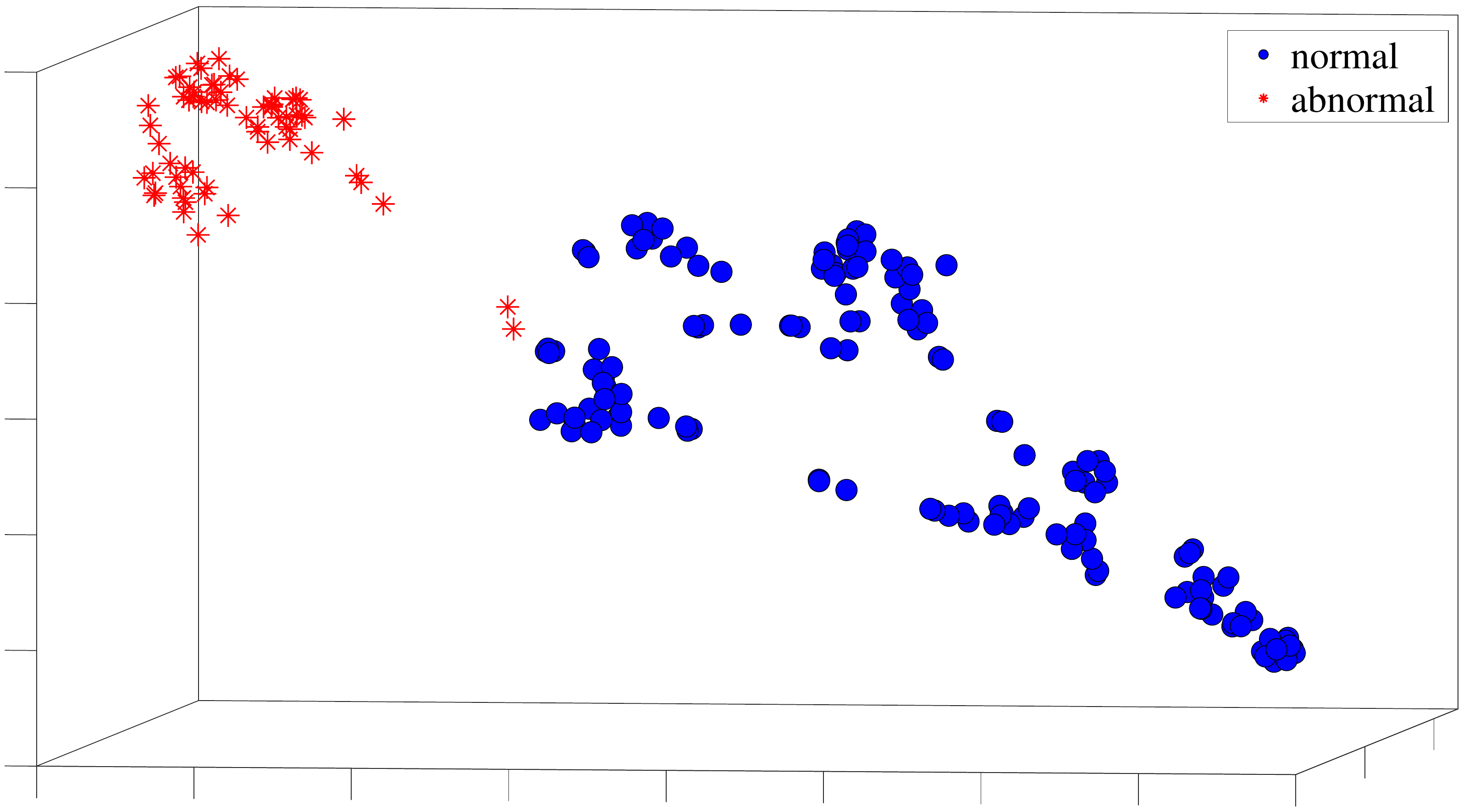}}		
	\subfigure[]{ \centering
		\label{data03}
		\includegraphics[width=0.308\textwidth]{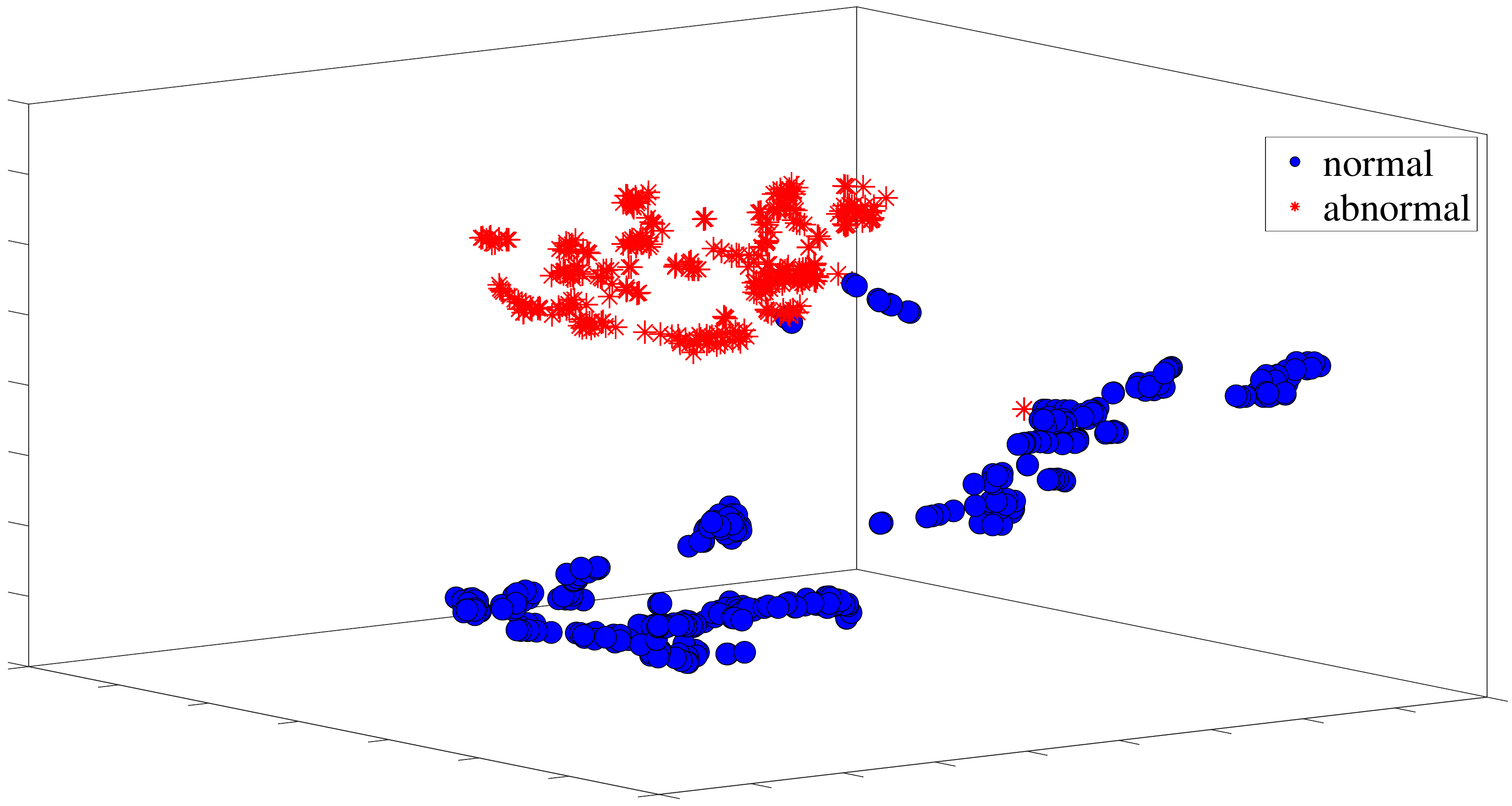}}
	\subfigure[]{ \centering
		\label{data09}
		\includegraphics[width=0.308\textwidth]{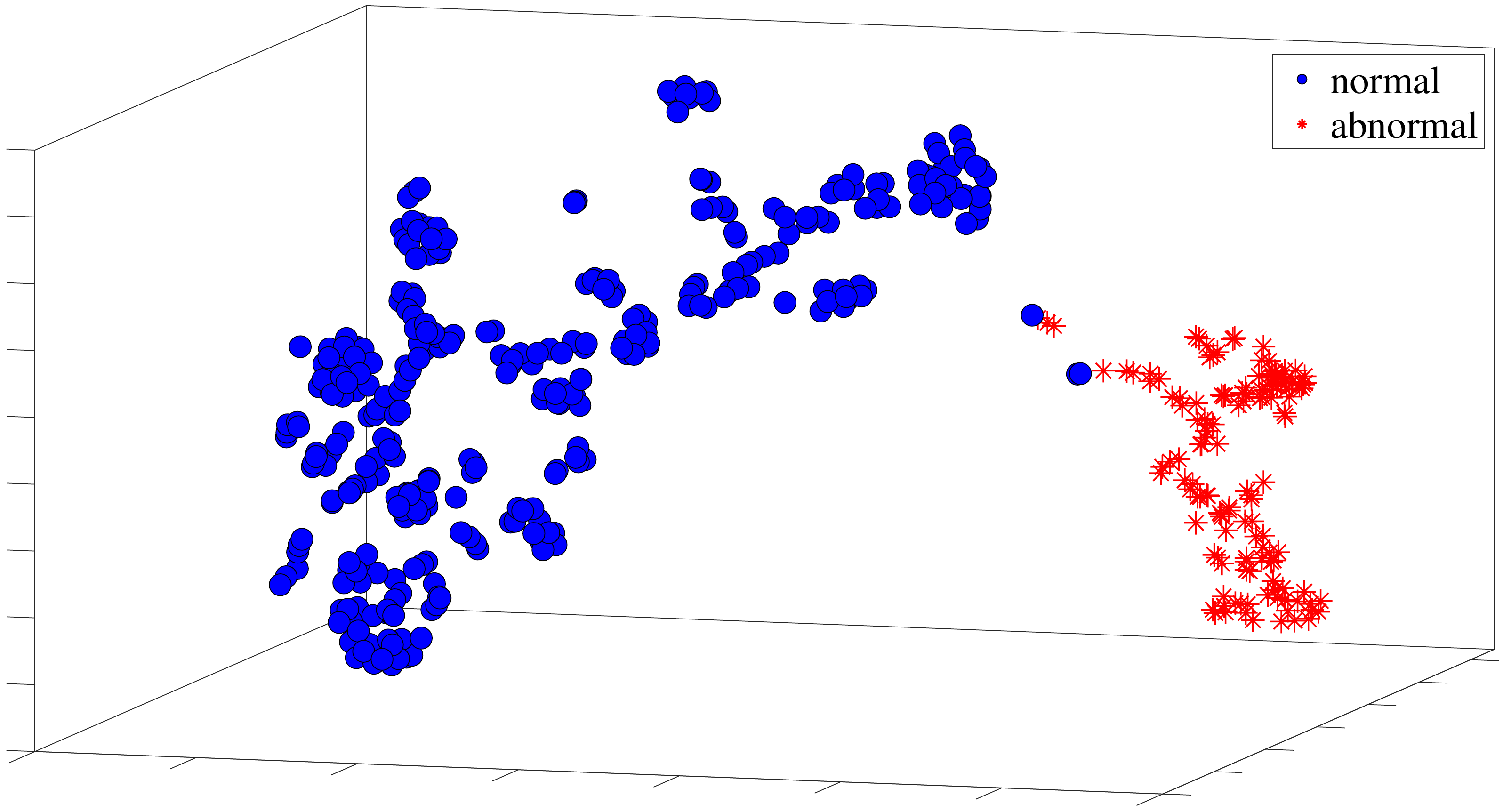}}
	\subfigure[]{ \centering
		\label{data091}
		\includegraphics[width=0.308\textwidth]{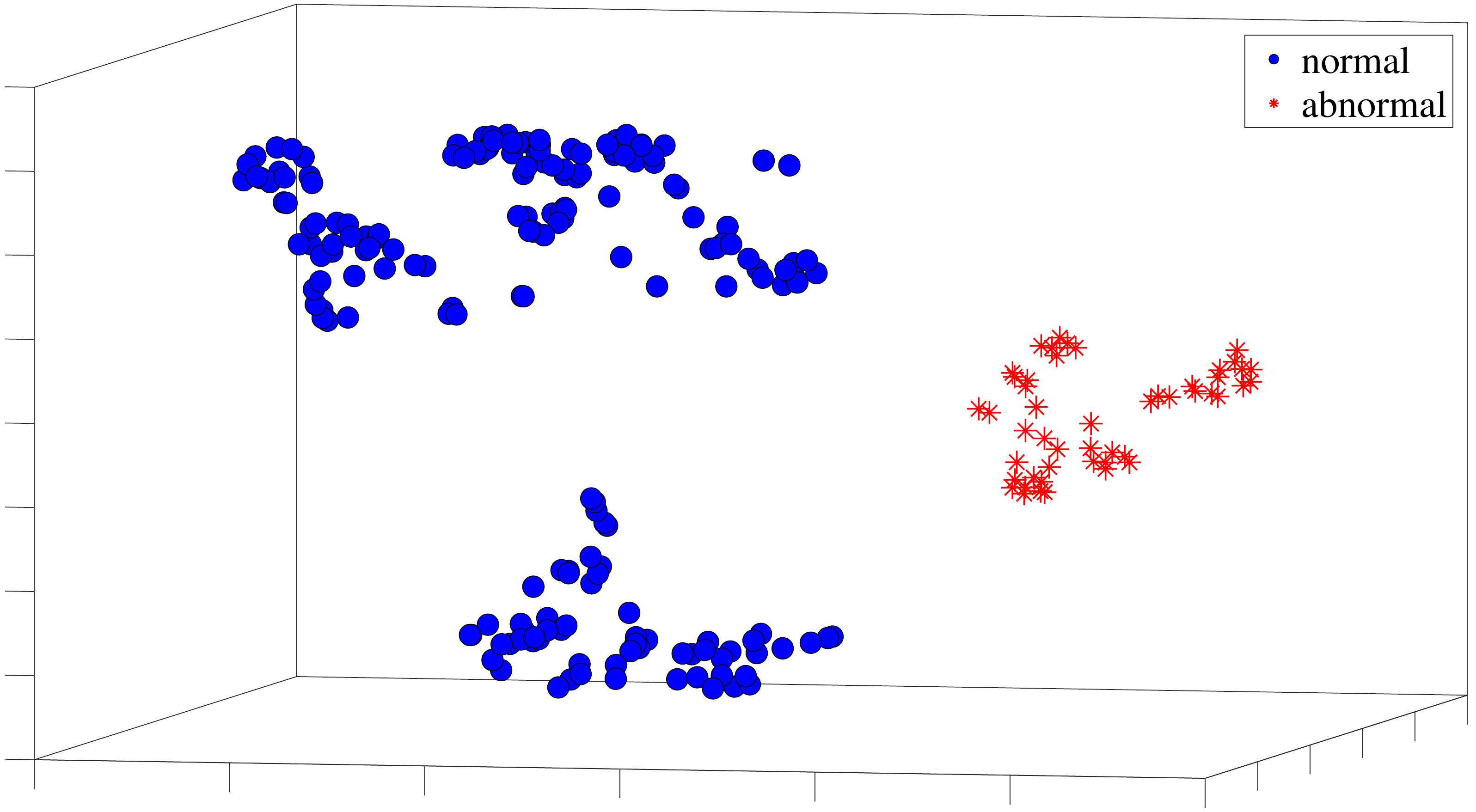}}
	\caption{Some results of visualization of the normal and abnormal data in the model space. It could be observed that in the model space, the model trained by the GPR images generated by the normal road is clearly separated from the model trained by GPR images generated from other underground structures. }
	\label{2class_result}
\end{figure*}

Fig. \ref{abexample} illustrates some identified abnormalities in the data.
Fig. \ref{2class_result} shows some results of OCSVM classification after mapping the GPR data obtained along the road from data space to model space. 
In order to intuitively show the classification effect of the model space, t-distributed Stochastic Neighbor Embedding (t-SNE) \cite{van2008visualizing} is used to reduce the obtained model to three dimensions for visualization. 
It should be pointed out that the actual model space is far more than three-dimensional. 
In the model space, it could be observed that the data is clearly divided into normal and abnormal by the blue and red points.

\begin{figure}[htpb]
	\centering
	\includegraphics[width=0.455\textwidth]{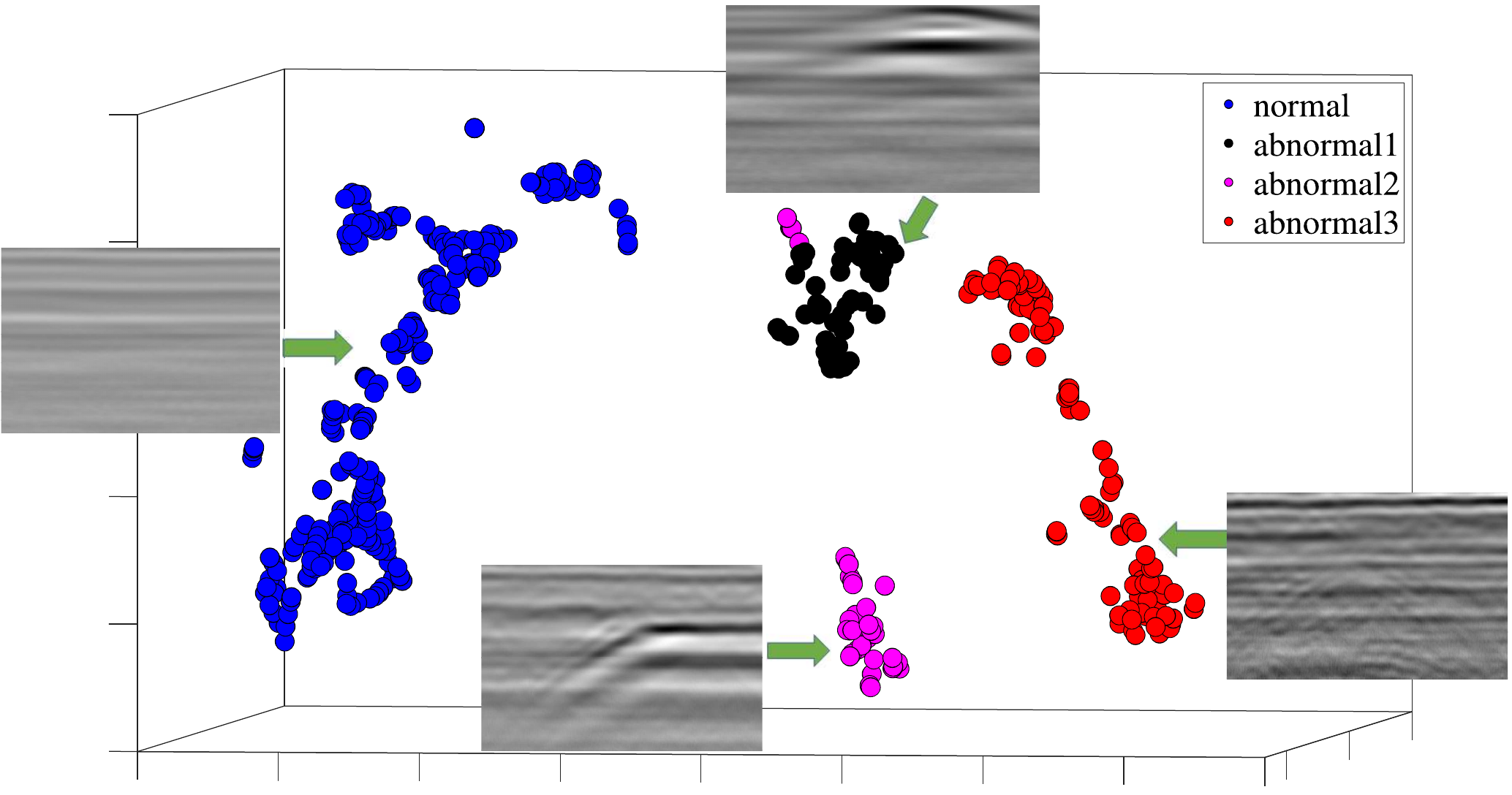}
	\caption{Visualization of the multi-abnormal dataset in the model space. In the model space, the window data is clearly divided into four categories representing normal, abnormal1, abnormal2, and abnormal3. Some sample GPR images are presented near each kind of the point. The blue points are normal data. Black and purple points indicate two different underground structures. The red points are also generated from the GPR data of a segment of the road. Different types of roads could also be identified in constructed model space. }
	\label{oneclass_result1}
\end{figure}

\begin{figure}[htbp]
	\centering
	\subfigure[]{ \centering
		\label{gpr1}
		\includegraphics[height=0.54in]{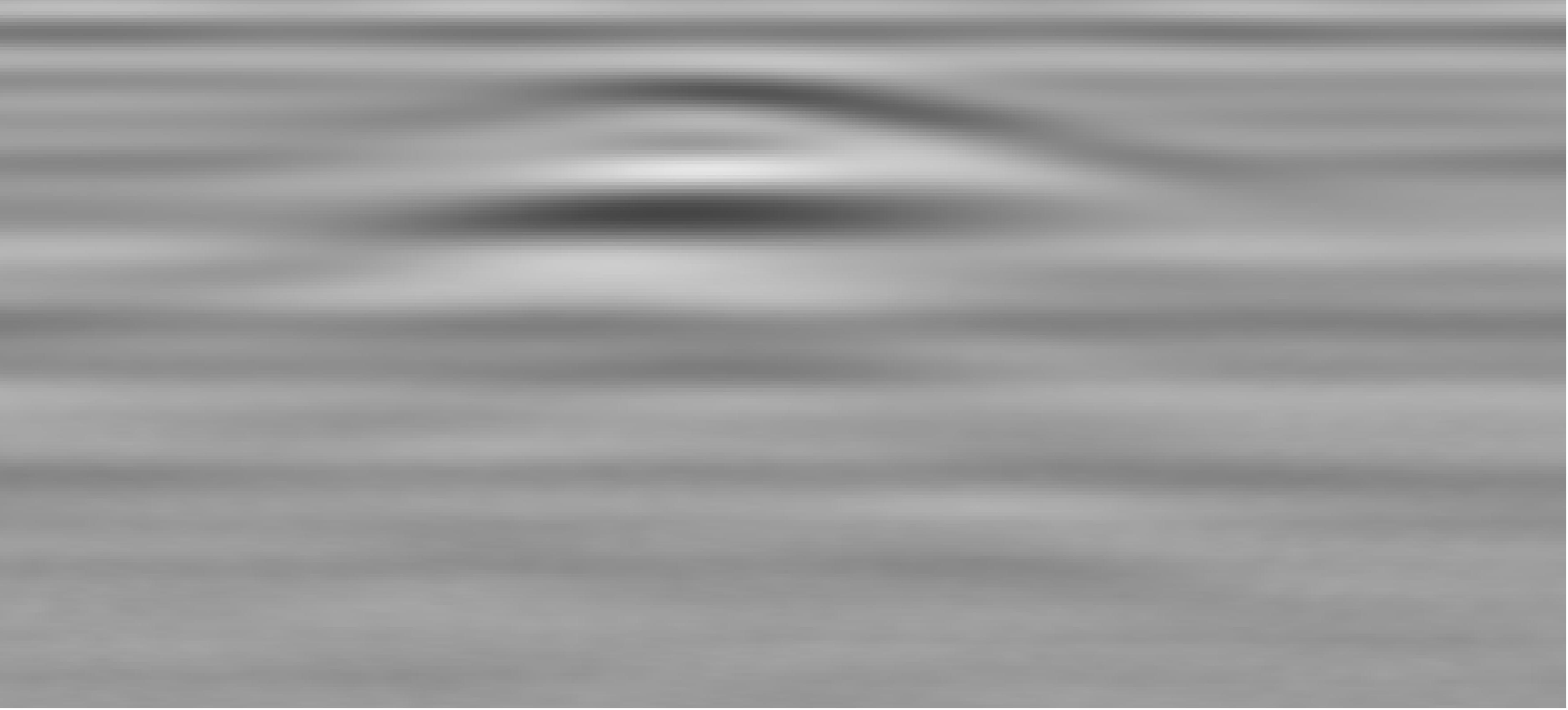}}
	\subfigure[]{ \centering
		\label{gpr3}
		\includegraphics[height=0.54in]{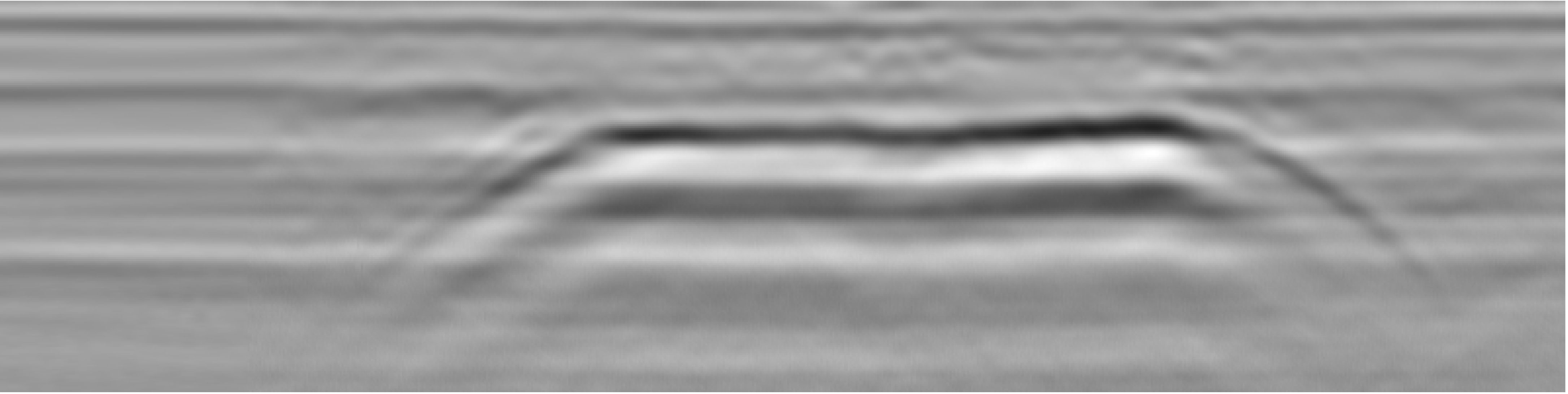}}
	\caption{These images show subsurface anomalies after merging the overlapping GPR image segments of the same type after incremental OCSVM. The results of the field inspection showed that (a) is an abandoned pipeline whose direction is not perpendicular to the detection direction of the GPR. (b) is an underground cavity.}
	\label{gprabnormal}
\end{figure}

As aforementioned, after performing OCSVM for the same type of abnormal data, we could further classify its mapped values in the model space through incremental OCSVM. A result of incremental OCSVM classification after mapping a GPR image from data space to model space with representative GPR images for each class is illustrated in Fig. \ref{oneclass_result1}. In our experiments, the definition of an abnormal is to define any other underground structures as abnormal data, except for the normal road underground situation without any subsurface facilities.
Intuitively, the GPR images of different underground structures are clustered into four categories in the model space. After completing the anomaly classification, the overlapping GPR image segments of the same type are merged to obtain the final anomaly GPR image since the length of some anomaly regions in the image exceeds the length of the sliding window. Fig. \ref{gprabnormal} shows some of the merged images. Specific results are presented in the following.

Since the collected data in this paper has been manually classified and some locations with part of underground anomalies were excavated and repaired. Supervised learning tasks could also be utilized in the conducted model space. In this case, we utilize the sliding window to generate a series of data segments. If there is no abnormal underground structure in the window, the window data is labeled as normal data. Otherwise, it is labeled as the corresponding abnormal data. In this experiment, we chose 8 GPR B-scan images obtained along 2 asphalt roads with similar underground environments, labeled and classify the underground abnormalities in 5 images to construct training sets and perform training. Then we use the well-trained model on the remaining 3 GPR images generated. Several representative supervised learning methods, including KNN \cite{altman1992introduction}, Random Forest \cite{ho1998random}, and SVM \cite{cortes1995support}, are utilized in the model space. 
Specific results of both semi-supervised and supervised learning tasks in the model space are presented in Tables \ref{ocsvm_result} and \ref{supervised_result}. 

\begin{table}[htbp]
	\centering
	\caption{The Result of Semi-Supervised Tasks in the 2D-ESN Model Space}
	\begin{tabular}{cccc}
		\toprule
		Road  & Precision & Recall & F1-score \\
		\midrule
		1     & $98.01\%$     & $100.00\%$     & $98.99\%$      \\
		\midrule
		2     & $97.36\%$     & $100.00\%$     & $98.66\%$      \\
		\midrule
		3     & $100.00\%$     & $97.65\%$     & $98.81\%$      \\
		\midrule
		4     & $100.00\%$     & $97.59\%$     & $98.78\%$      \\
		\bottomrule
	\end{tabular}%
	\label{ocsvm_result}%
\end{table}%

\begin{table}[htbp]
	\small
	\centering
	\caption{The Result of Supervised Tasks in the 2D-ESN Model Space}
	\begin{tabular}{cccc}
		\toprule
		Algorithm & Precision & Recall & F1-score  \\
		\midrule
		KNN   & $\mathbf{97.74\%}$     & $\mathbf{98.85\%}$     & $\mathbf{98.29\%}$     \\
		\midrule
		Random Forest & $97.31\%$     & $96.79\%$     & $97.04\%$     \\
		\midrule
		SVM   & $93.70\%$     & $96.57\%$     & $95.11\%$     \\
		\bottomrule
	\end{tabular}%
	\label{supervised_result}%
\end{table}%

\begin{figure}[htbp]
	\centering
	\subfigure[]{ \centering
		\label{boundry1}
		\includegraphics[width=0.49\textwidth]{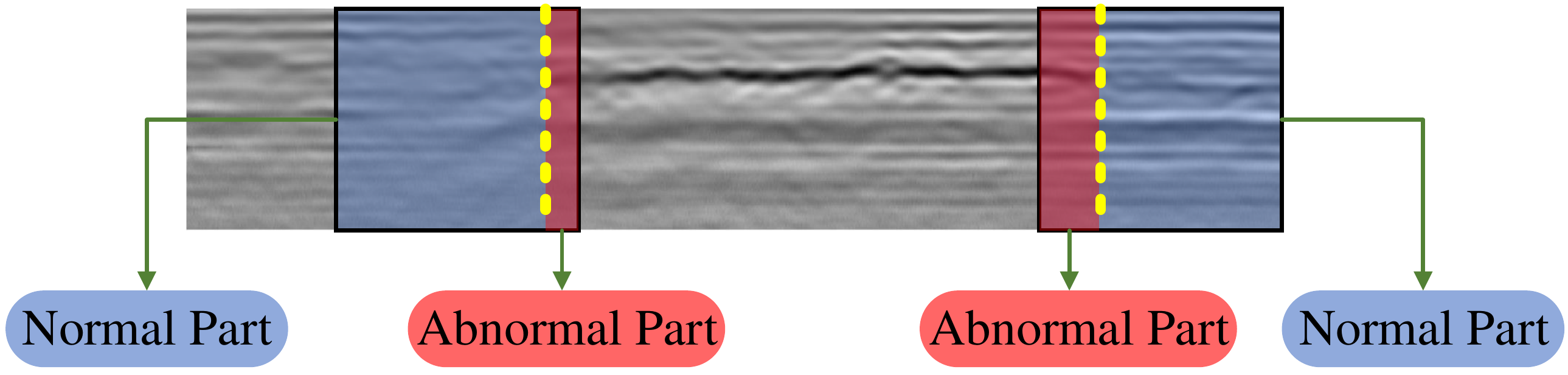}}
	\subfigure[]{ \centering
		\label{boundry2}
		\includegraphics[width=0.49\textwidth]{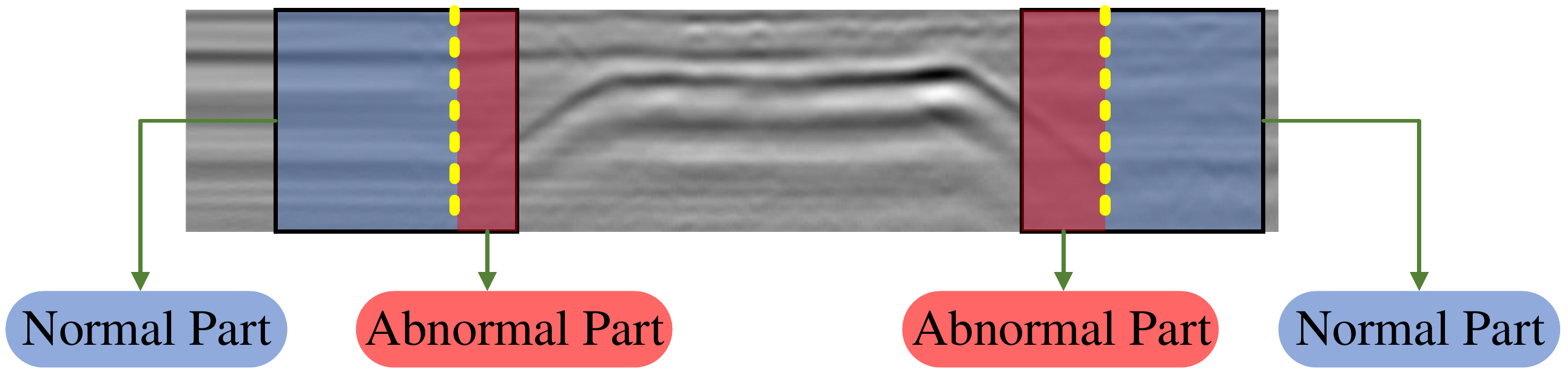}}
	\caption{Schematic diagram of transition stage of GPR image. (a) and (b) are two examples of underground abnormals, in which the blue part of the sliding window is the normal part, the red part is the abnormal part, and the yellow dotted line is the dividing line. The GPR image state transition data shown here are mapped to transition points in the model space.}
	\label{boundry}
\end{figure}

\begin{figure}[htpb]
	\centering
	\includegraphics[width=0.45\textwidth]{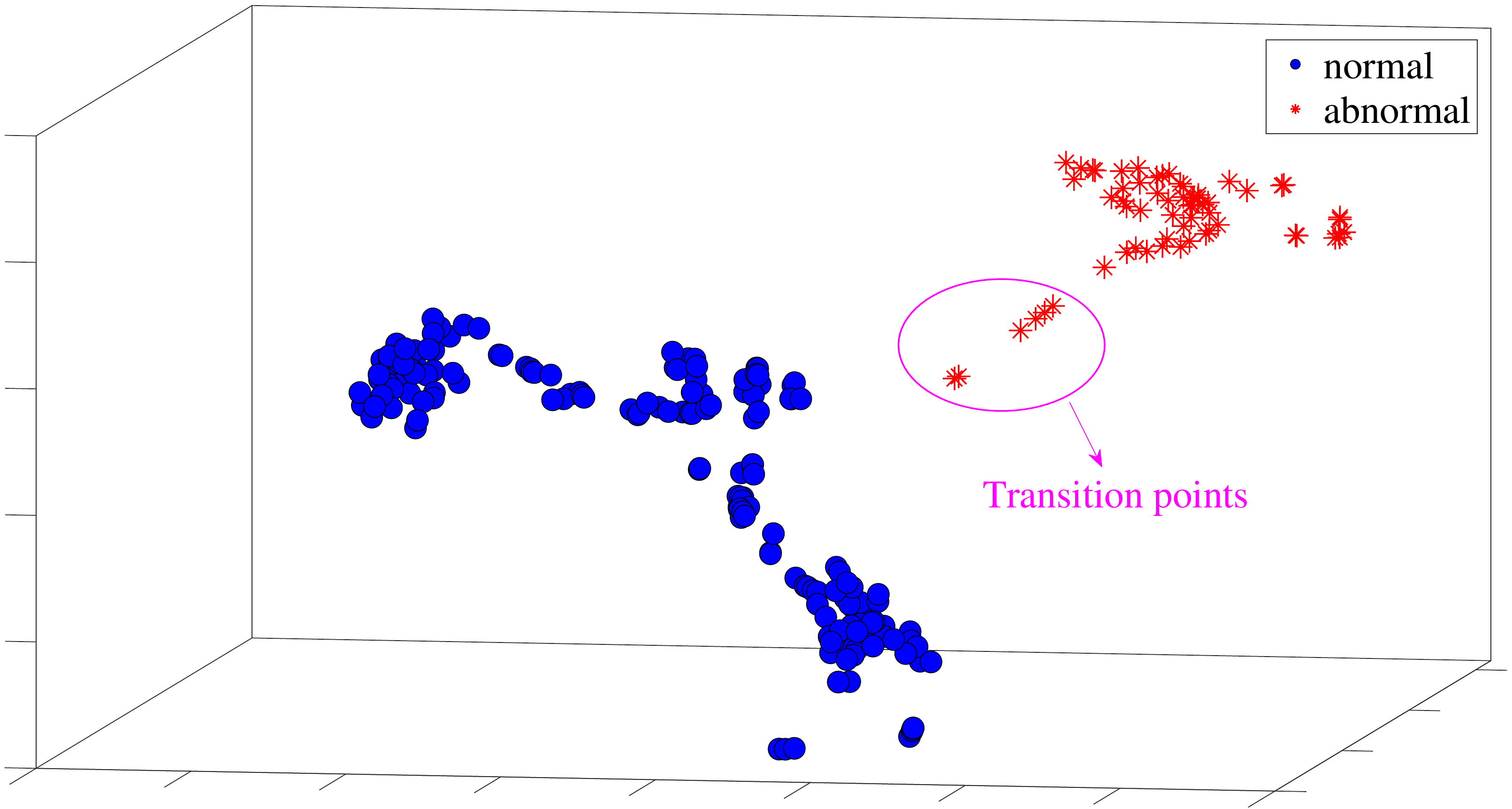}
	\caption{Visualization of the transition points in the model space. In the model space, the points in the magenta circle are the part where the sliding window starts to slide to the abnormal part. }
	\label{data09_2_analyse}
\end{figure}

In our experiments, no matter supervised or unsupervised learning, the method proposed in this paper does not require many or multiple types of data for training. In terms of time, the method in this paper only takes 0.04 seconds to process each sliding window.
The wrong classification mainly comes from the window data that has just stepped from the normal part to the abnormal part or is about to leave the abnormal part, as shown in Fig. \ref{boundry}.
The data at this kind of window would be mapped to ``transition points'' in the model space as shown in Fig. \ref{data09_2_analyse}. 
These transition points reflect the sensitivity of the model space method to anomalies, but to some extent, this phenomenon would cause confusion when sliding through two or more adjacent anomalies. Therefore, further studies could be performed on the trajectories of points in the model space to improve the accuracy of anomaly classification. On the other hand, in real-world applications, the positioning data could be erroneous. In the case of tall buildings and trees, satellite-based positioning signals could be blocked \cite{li2014estimating}. The positioning accuracy of an odometer would degrade when a rough or slippery ground is measured\cite{prokhorenko2012topographic}.
Thus using a range slightly larger than the anomaly area to locate the subsurface anomaly could improve the robustness of anomaly detection.

\subsection{Comparison and Analysis}

To evaluate the efficiency of the proposed method, the obtained GPR B-scan images are processed by the Histograms of Oriented Gradient (HOG)\cite{dalal2005histograms}, deep residual learning (ResNet)\cite{he2016deep}, and AlexsNet\cite{krizhevsky2012imagenet}. The four methods (HOG+SVM, AlexsNet, ResNet-18, and KNN in the constructed model space) could obtain the location of abnormal structures in the GPR B-scan image and classify anomalies. Specific results of the comparison are presented in Table \ref{comparsion}. 

\begin{table}[htbp]
	\small
	\centering
	\caption{Comparison Results on HOG, AlexNet, ResNet18, and the proposed method}
	\begin{tabular}{ccccc}
		\toprule
		Algorithm & Precision & Recall & F1-score & Time(s)\\
		\midrule
		HOG+SVM   & $98.77\%$     & $92.37\%$     & $95.46\%$     & 2.57 \\
		\midrule
		AlexNet  & $95.01\%$     & $96.18\%$     & $95.55\%$     & 417 \\
		\midrule
		ResNet-18   & $\mathbf{98.98\%}$     & $92.94\%$     & $95.87\%$     & 543 \\
		\midrule
		Proposed Method  & $97.74\%$     & $\mathbf{98.85\%}$     & $\mathbf{98.29\%}$     & 20.71 \\
		\bottomrule
	\end{tabular}%
	\label{comparsion}%
\end{table}%
\begin{figure}[htbp]
	\centering
	\subfigure[]{ \centering
		\label{model_05}
		\includegraphics[width=0.36\textwidth]{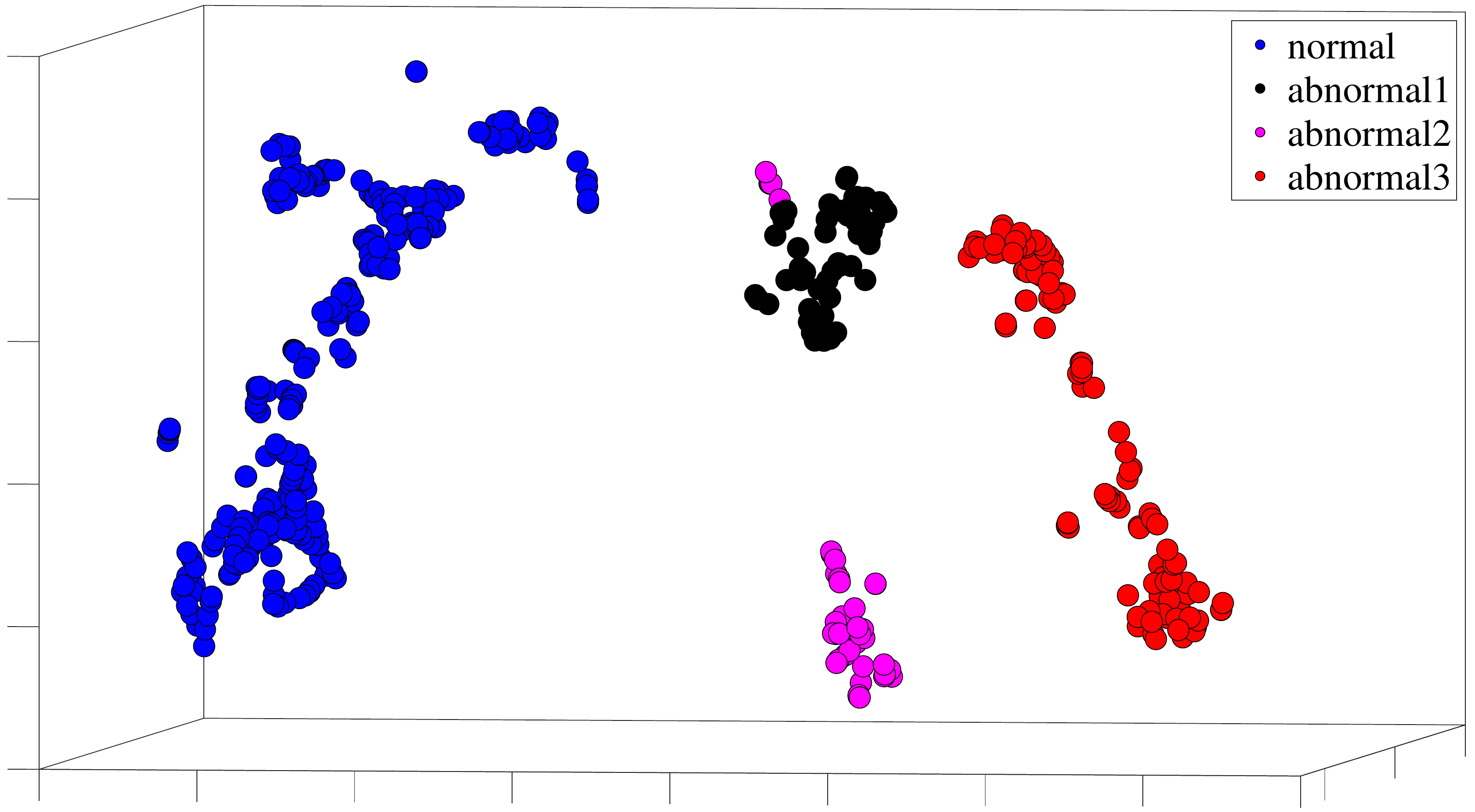}}
	\subfigure[]{ \centering
		\label{hog_05}
		\includegraphics[width=0.36\textwidth]{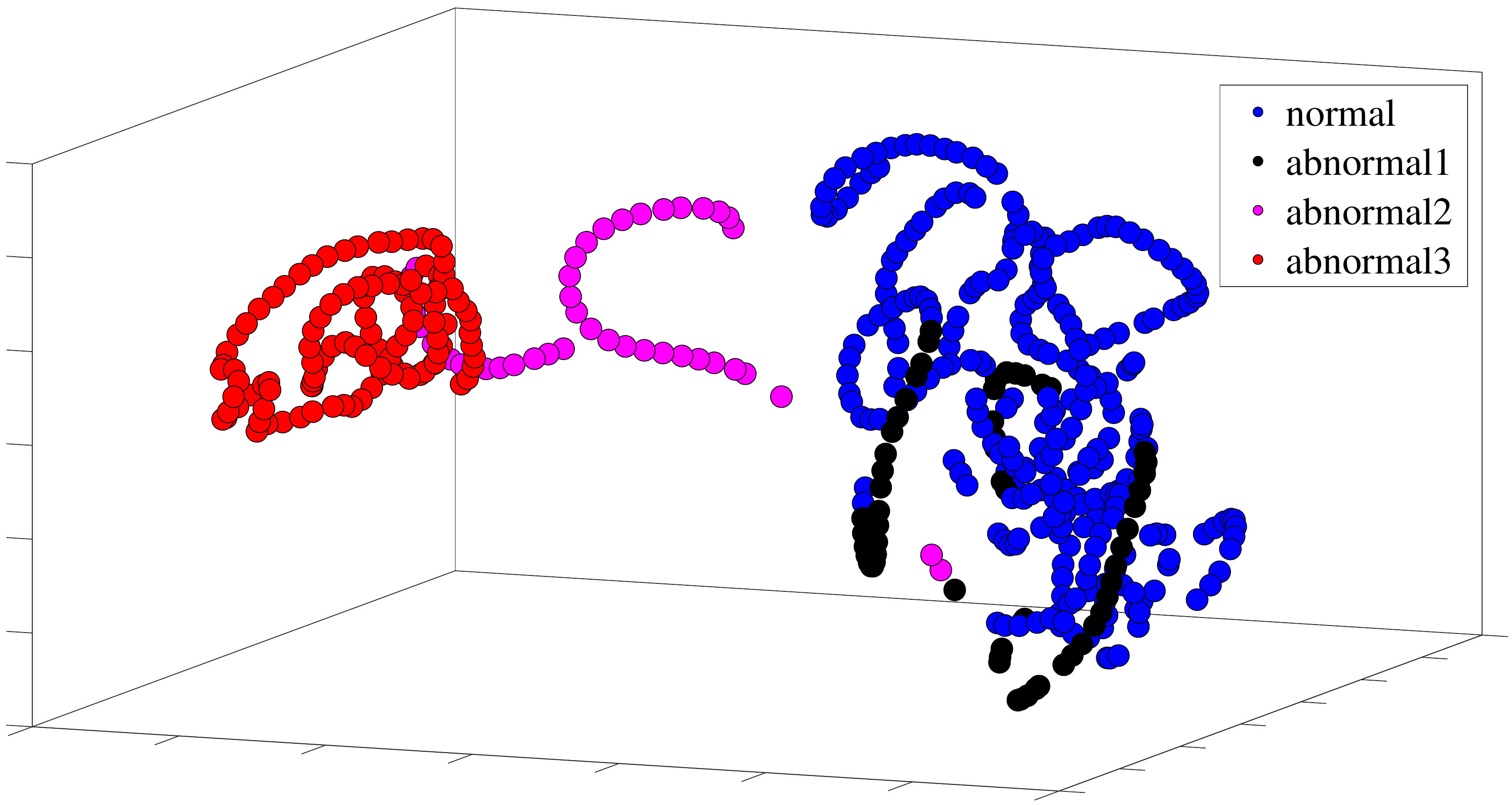}}
	\caption{Visualization of the window data in the (a) model space and (b) HOG feature space. Compared with the HOG feature space, in the model space, the data has a better clustering result.}
	\label{2method_result}
\end{figure}

Histogram of Oriented Gradient (HOG) is a feature descriptor used in image processing for object detection. This algorithm counts the number of occurrences of gradient directions in local parts of the image. In this paper, HOG with the cell of size $32\times32$ is utilized to extract the feature of the window data. And then, SVM is used for classification in the HOG feature space. The visual comparison between HOG feature space and model space is shown in Fig. \ref{2method_result}. It could be observed that the blue and black points in HOG are intertwined and difficult to separate. Intuitively, different types of GPR data in the model space have larger inter-class distances and are easier to classify. As shown in Table \ref{comparsion}, as a traditional feature extraction algorithm, HOG has high computational efficiency. However, HOG+SVM achieves the lowest recall, which is the most important metric in underground anomaly diagnosis.

AlexNet and ResNet-18 both are deep convolutional neural networks (CNN), which require pre-training to identify and classify tasks. From the results, although CNN can realize the identification of underground anomalies and its accuracy and other indicators are similar to those of the method proposed in this paper, its training time is much longer than that of the proposed method based on learning in the model space (including semi-supervised and supervised learning in the model space). On the other hand, the CNN method needs to try to ensure that the training set is similar to the underground environment of the test set. We used the above-trained model to identify and classify abnormal structures in the GPR images of the other 2 roads. Due to changes in the underground environment, the classification accuracy dropped significantly to about 80\%. By mapping the GPR image into the model space, since the distance between different model points reflects the difference in the data generation model, that is 2D-ESN, the metric information between the models helps to mine the essential characteristics of the data. In the conducted experiments, both supervised learning and semi-supervised learning in the model space have achieved better results than those in the data space.

From the results of the above experiments and comparisons, the method in this paper has achieved considerable results. The proposed 2D-ESN essentially captures the dynamic characteristics of GPR images by taking into account correlations in two directions, which is one of the reasons why clustering algorithms in the model space get accurate. In fact, not only GPR B-scan images but also data that have continuity in two or more directions, such as computed tomography (CT) data,  have this kind of feature. Since these data or images do not belong to the scope of the underground anomaly diagnosis described in this paper, they will not be discussed and analyzed here. In the following work, we will continue to study and expand the method based on learning in the model space and hope to build different fitting models for different kinds of data.

\section{Conclusion}

In this paper, a GPR B-scan image diagnosis method based on learning in the model space is proposed. A sliding window is constructed and swiped across the obtained GPR image. The GPR image in each sliding window is mapped into model space by the proposed 2D-ESN. The distance measuring method for 2D-ESN is modified in the constructed model space. Based on the constructed model space and distance measuring method, supervised and semi-supervised algorithms could be utilized to identify or classify anomalies on GPR B-scan images. Both detection position and time continuity in the horizontal direction and medium continuity in the vertical direction of the GPR B-scan image are taken into account in the proposed 2D-ESN. Therefore, the proposed 2D-ESN processes images from the point level, which can effectively reduce the dimension of the fitted model, thereby reducing the memory usage. When applying the proposed method, only a normal GPR image segment of the detection area (i.e., no subterranean anomalies) is required to start the diagnosis.
Experiments on real-world datasets are conducted, and both supervised and semi-supervised learning in the model space have achieved considerable results. In future work, we plan to extend 2D GPR B-scan images to multi-dimensional correlated data and design models for these data to achieve efficient mapping from data space to model space. On the other hand, we will study the model change trajectory in the model space (that is, the trajectory of the point in the model space) and design abnormality warning algorithms.

\bibliography{ref}
\bibliographystyle{IEEEtran}

\end{document}